\documentclass[letterpaper, 10 pt, conference]{ieeeconf}  

\IEEEoverridecommandlockouts                              

\usepackage{url}
\usepackage{cite}
\usepackage{amsmath,amssymb,amsfonts}
\usepackage{algorithm}
\usepackage{algorithmic}
\usepackage{graphicx}
\usepackage{textcomp}
\usepackage{xcolor}
\usepackage{bm}
\usepackage{svg}

\newcommand{\rep}{\text{rep}}

\usepackage{subcaption} 
\usepackage[colorinlistoftodos]{todonotes} 

\title{\LARGE \bf 3D Reactive Control and Frontier-Based Exploration for Unstructured Environments \\
\thanks{$^{1}$ Department of Aerospace Engineering Sciences, University of Colorado Boulder, CO, \{shakeeb.ahmad, andrew.b.mills, eric.frew\}@colorado.edu}
\thanks{$^{2}$ Department of Mechanical Engineering, University of Colorado Boulder, CO, \{eugene.rush, sean.humbert\}@colorado.edu}
\thanks{$^{*}$ These authors contributed equally to this work}
\thanks{This work was supported through the DARPA Subterranean Challenge, cooperative agreement number HR0011-18-2-0043}}

\author{Shakeeb Ahmad$^{1*}$, Andrew B. Mills$^{1*}$, Eugene R. Rush$^{2*}$, Eric W. Frew$^{1}$, J. Sean Humbert$^{2}$}

\begin{document}

\maketitle
\thispagestyle{empty}
\pagestyle{empty}

\begin{abstract}
The paper proposes a reliable and robust planning solution to the long range robotic navigation problem in extremely cluttered environments. A two-layer planning architecture is proposed that leverages both the environment map and the direct depth sensor information to ensure maximal information gain out of the onboard sensors. A frontier-based pose sampling technique is used with a fast marching cost-to-go calculation to select a goal pose and plan a path to maximize robot exploration rate. An artificial potential function approach, relying on direct depth measurements, enables the robot to follow the path while simultaneously avoiding small scene obstacles that are not captured in the map due to mapping and localization uncertainties. We demonstrate the feasibility and robustness of the proposed approach through field deployments in a structurally complex warehouse using a micro-aerial vehicle (MAV) with all the sensing and computations performed onboard.
\end{abstract}

\section{Introduction}
Robotic exploration and navigation through unknown and unstructured environments have many real-world applications, ranging from firefighting and surveillance to search and rescue operations for natural disasters. A successful autonomous robotic mission may result in very valuable information about a previously unknown environment that could be hazardous for humans. A recent DARPA Subterranean (SubT) Challenge \cite{subt} poses such a problem for large scale, complex underground environments. A large variety of these environments require reliable and repeatable solutions to 3D navigation and planning. A popular vehicle choice is the quadrotor UAV due to its traversability and hover-in-place capabilities. This paper scopes 3D exploration and planning for a quadrotor UAV in unknown environments with extreme levels of clutter. \par

\begin{figure}
    \centering
    \includegraphics[width=1.0\linewidth]{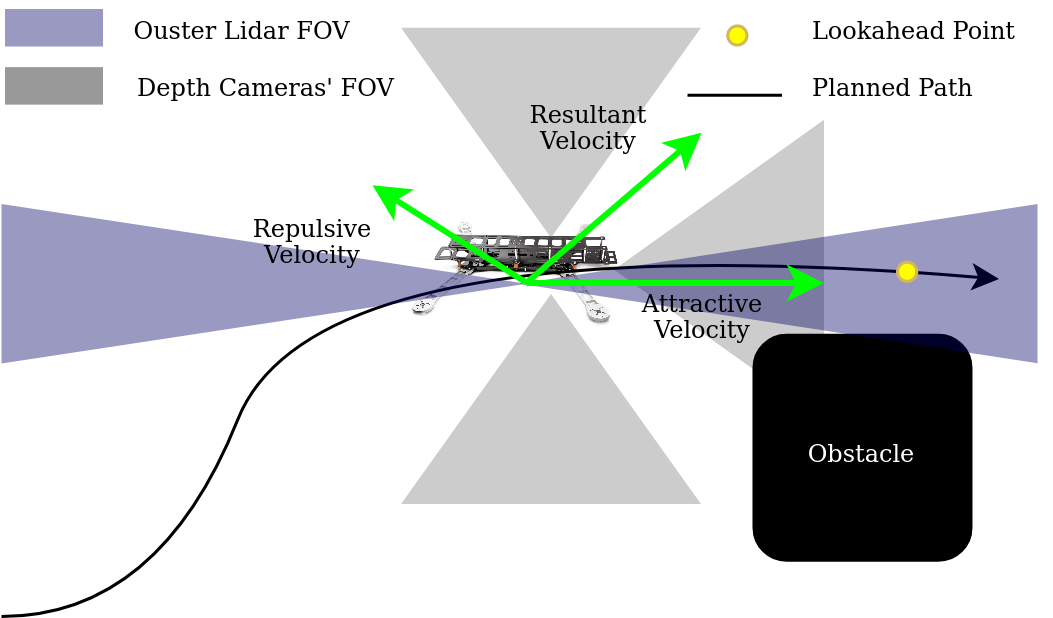}
    \caption{A side view of the depth sensors' configuration on the quadrotor UAV. The environment map is incrementally built using the upward and downward facing depth cameras and the long range 360$^{\circ}$ Ouster lidar. The path is generated using the map-based frontier view planner and followed by the APF reactive controller which uses direct information from all the depth cameras to locally avoid small obstacles.}
    \label{fig:sensor_config}
\end{figure}

Typical motion planners rely on one of the two types of information: instantaneous sensor measurements or a map representation of the environment. The map is typically built in an incremental fashion as the vehicle traverses through an unknown environment by fusing the environment information as seen by the robot. OctoMap \cite{hornung13octomap} and Voxblox \cite{oleynikova2017voxblox} have been shown in past to provide dense representation of an environment in a computationally feasible way. Given the occupancy grid of an area already explored by the robot, Yamauchi \cite{yamauchi1997frontier,yamauchi1998mobile} first proposed the concept of frontier-based exploration. He proposed that efficient exploration of unknown areas could be achieved by detecting map frontiers, grouping them together and prioritizing them based on a user-defined objective. He further presented \cite{yamauchi1998frontier} that the technique had a potential to work in a multi-robot setting where each robot could share map frontiers making the exploration resilient to the loss of one or multiple robots in the network. The concept of frontiers has been widely used since then to bias path planners towards the unseen areas in a map \cite{freda2005frontier,gonzalez2002navigation, heng2015efficient}. Our path planning method is primarily based on this notion. The environment representation is obtained in the form of Euclidean Signed Distance Field (ESDF), a concept that is well-known in computer graphics and has been proven very valuable to many path planners in past especially in mobile robotics \cite{dharmadhikari2020motion,russell2020learning}. We use Voxblox as an open-source and computationally feasible solution to obtain this information. Frontiers are first grouped together to obtain multiple goal poses of interest at any replan time instant. Fast marching method is then used to calculate the appropriate cost-to-go from the robot's position to the sampled goal poses and the lookahead path is generated by following the gradient of the cost-to-go map.   \par

In order to ensure reliable navigation through a highly cluttered environment with obstacles of heterogeneous nature, a planner relying on a single type of information may not be sufficient. This is because the map-based motion plans rely heavily on the mapping accuracy. Small and thin objects are typically difficult to capture in maps due to their reduced representation. Building a very high resolution map is computationally costly for a small-scale robot such as an MAV. Moreover, small perturbations in the onboard localization add to the mapping uncertainty. As a result, the motion plans may get extremely close to the obstacles in general or may ignore small scene obstacles completely that are non-existent in the map, rendering the solution ineffective especially for a UAV that is highly sensitive to collisions. The reactive controllers inspired by potential functions \cite{khatib1986real} and biology \cite{conroy2009implementation} solve many of these issues by planning relative to the sensor frame without building an environment map. Some efforts have been made recently to plan inside the depth image space \cite{ahmad2021apf, fragoso2018dynamically, dubey2017droan, matthies2014stereo, ahmad2019real}. We complement the mapping-based planning solution with a potential function reactive controller that senses and avoids small obstacles using direct depth information from the onboard sensors. In the rest of our paper, we use the terms \textit{global} and \textit{local} for the map-based planner and the reactive controller respectively. Existing long-range UAV planning methods in the literature such as \cite{bircher2016receding,dang2019explore,Dang2020,dharmadhikari2020motion} have been demonstrated in underground tunnels and small confined nearly-empty rooms. These environments exhibit general predictable structures, for instance, underground mines typically consist of long corridors connected through junctions with most of the obstacles appearing close to the walls. We demonstrate our planning approach in a highly cluttered and unstructured environment containing obstacles that are heterogeneous in shape and size. In one of our runs, the UAV travelled over 300 m in a tightly-spaced environment. To the best of our knowledge, this work is the first attempt to demonstrate the long-range UAV exploration in the environments of extreme clutter. The solution is a part of the autonomy stack of our team (MARBLE) for the DARPA Subterranean Challenge Phases I and II \cite{ohradzansky2021mutiagent}. \par

The rest of the paper is organized as follows. Section \ref{sec:problem_setup} motivates the problem and introduces the sensing modalities. Section \ref{sec:global_planning} details the map-based path planner followed by the discussion of the vision-based reactive controller in Section \ref{sec:local_planning}. The results from the physical UAV experiments are presented in Section \ref{sec:experimental_evaluation}. Finally, Section \ref{sec:conclusion} concludes the discussion.

\section{Problem Setup}
\label{sec:problem_setup}

In order to maximize the field-of-view (FOV) in a practically feasible manner, the UAV considered in this paper is equipped with 4 depth sensors: a 360 degrees long-range \textit{3D lidar sensor} and upward, downward and forward facing \textit{depth cameras}. The 3D lidar sensor used for our application is Ouster OS1 which is well-suited for localization and mapping due to its long range and wide horizontal FOV. However, like many other long range lidar sensors, it suffers from narrow vertical FOV, sparse pointcloud, and performance degradation at close ranges. In contrast, the depth cameras, such as an Intel Realsense, provide a higher resolution depth information and wider vertical FOV, that are essential to detect and avoid small and thin obstacles at close ranges. Therefore, the local obstacle avoidance relies only on the upward, downward and forward facing depth cameras while the map for the global planner is built primarily using the 3D lidar sensor complemented by the upward and downward facing depth cameras for vertical awareness. In order to compensate for the cameras' horizontal FOV, we consider the UAV to be exhibiting the nonholonomic unicycle kinematics in the $x-y$ plane of its body frame $\mathcal{F}_{\mathcal{B}}$ where $v_x$, $v_z$, and $v_\theta$ are the forward, vertical and steering command velocities in $\mathcal{F}_{\mathcal{B}}$ respectively. This ensures that there is a depth camera in every allowed direction of motion for the local obstacle avoidance. The pose of the robot in the world frame $\mathcal{F}_{\mathcal{W}}$ is referred to as $(x,y,z,\theta) \in \mathcal{X} \subset \mathbb{R}^3 \times S^1$ where $\mathcal{X}$ is the set of all possible robot poses. The outer planning loop generates a lookahead path at each replan time instant using the most recent environment map available. With an intention to plan in a computationally feasible manner onboard an MAV with limited resources, the planned \textit{lookahead path} is purely geometric. The inner loop is primarily based on the Artificial Potential Function (APF) controller and is responsible for generating steering and translational velocity commands for the robot, given a lookahead path. In order to follow the path, we choose a \textit{lookahead point} on the path that is a constant distance away from the robot at all times. This point serves as the attractor for the APF controller and moves along the path as the robot navigates, until the next replan time instant or the end of path. Since the controller directly generates the control commands, $v_x$, $v_z$, and $v_\theta$, the nonholonomic constraints are taken care of during path following. The controller relies on the direct sensor information from all the depth cameras in order to generate the repulsive velocity vector. This two-tier collision avoidance \textit{i.e.}, map-based and map-less, plays an important role in improving the navigation safety in practice by dealing with the uncertainties that arise due to imperfect path following coupled with the mapping and localization uncertainties during path planning. The next two sections provide further insight into the two planning layers.

\section{Global Path Planning}
\label{sec:global_planning}

\begin{algorithm}
\begin{algorithmic}[1]
  \STATE path$ \gets \{\}$ \\
  \STATE $\mathcal{F} \gets $FindFrontierVoxels$(\mathcal{M})$
  \STATE $\mathcal{F}_{cl} \gets$FilterFrontierContiguous$(\mathcal{F},N_{c})$\\
  \STATE $\mathcal{P} \gets $SampleViewPoses$(\mathcal{F}_{cl},r_{g})$
  \STATE $T \gets $ComputeFMM$(\mathcal{M},c(\textbf{x}))$ \\
  \STATE $\textbf{p}_{goal} \gets \max_{\textbf{p} \in \mathcal{P}} U(\textbf{p}, \textbf{x})$ \\
  \STATE path$ \gets $FollowGradient$(c(\textbf{p}_{goal}), v(\textbf{x}))$ \\
\end{algorithmic}
\caption{Planner$(\mathcal{M},\textbf{x}, N_{c}, r_{g})$}
\label{algorithm:FrontierPlanning}
\end{algorithm}

\subsection{Preliminaries}
It is assumed that the robot is able to actively build and maintain a consistent 3D voxelized grid cell representation of the environment using onboard state estimation and depth sensors.  The map $\mathcal{M}$ is a set of voxelized grid cells $\bm{c}$, defined as 
\begin{equation}
    \mathcal{M} = \{ \bm{c} : \bm{c} \in \mathbb{Z}^3,
    \bm{c} \succeq \bm{0},
    \bm{c} \preceq \bm{c}_{\text{max}.}
    \}
\end{equation}
Every voxel in the map corresponds to a physical position in the real environment.  This position transform is defined as
\begin{equation}
    p(\bm{c}) = \bm{c}s_c + \bm{d}_m \in \mathbb{R}^3
\end{equation}
where $p(\bm{c})$ is the position of the voxel, $s_c$ is the voxel size, and $\bm{d}_m$ is the map datum position.  

Two functions are defined at each voxel position $\bm{c}$. The first function is the voxel occupancy probability $O(\bm{c}): \mathcal{M} \rightarrow [0,1]$ with $0$ and $1$ referring to the voxel as free and occupied respectively. Unseen voxels begin with an occupancy probability of $0.5$.  The second is the distance to the closest occupied cell $D(\bm{c}): \mathcal{M} \rightarrow \mathbb{R}$.  This is a Euclidean Signed Distance Field (ESDF) where negative values of $D(\bm{c})$ are the distance to the closest free cell when $\bm{c}$ is occupied. In practice, we use Voxblox to maintain the ESDF \cite{oleynikova2017voxblox}.

For the purpose of global planning, we consider only positions in $\mathbb{R}^3$ for online computation.  
\begin{equation}
    \bm{r} = \begin{bmatrix}
    x \quad y \quad z
    \end{bmatrix}^T \in \mathbb{R}^3
\label{eq_state}
\end{equation}
The space of safe robot configurations restricts the geometric position of the robot to all voxel cells that are at least $d_{safe}$ away from the nearest occupied voxel.  The safe voxel set, $\mathcal{X}_{safe}$ is defined as

\begin{equation}
\begin{aligned}
    \mathcal{X}_{safe} &= \{ \bm{x} : c(\bm{x}) \in \mathcal{M}, D(c(\bm{x})) \geq d_{safe} \} \\
    c(\bm{r}) &= \text{round} \bigg( \frac{\bm{r} - \bm{d}_m}{s_c} \bigg) \in \mathbb{Z}^3
\end{aligned}
\label{eq_state_configuration}
\end{equation}
where $c(\bm{r})$ maps from the robot position in $\mathbb{R}^3$ to the nearest map voxel $\in \mathbb{Z}^3$.

\begin{figure}
    \centering
    \includegraphics[width=0.49\textwidth]{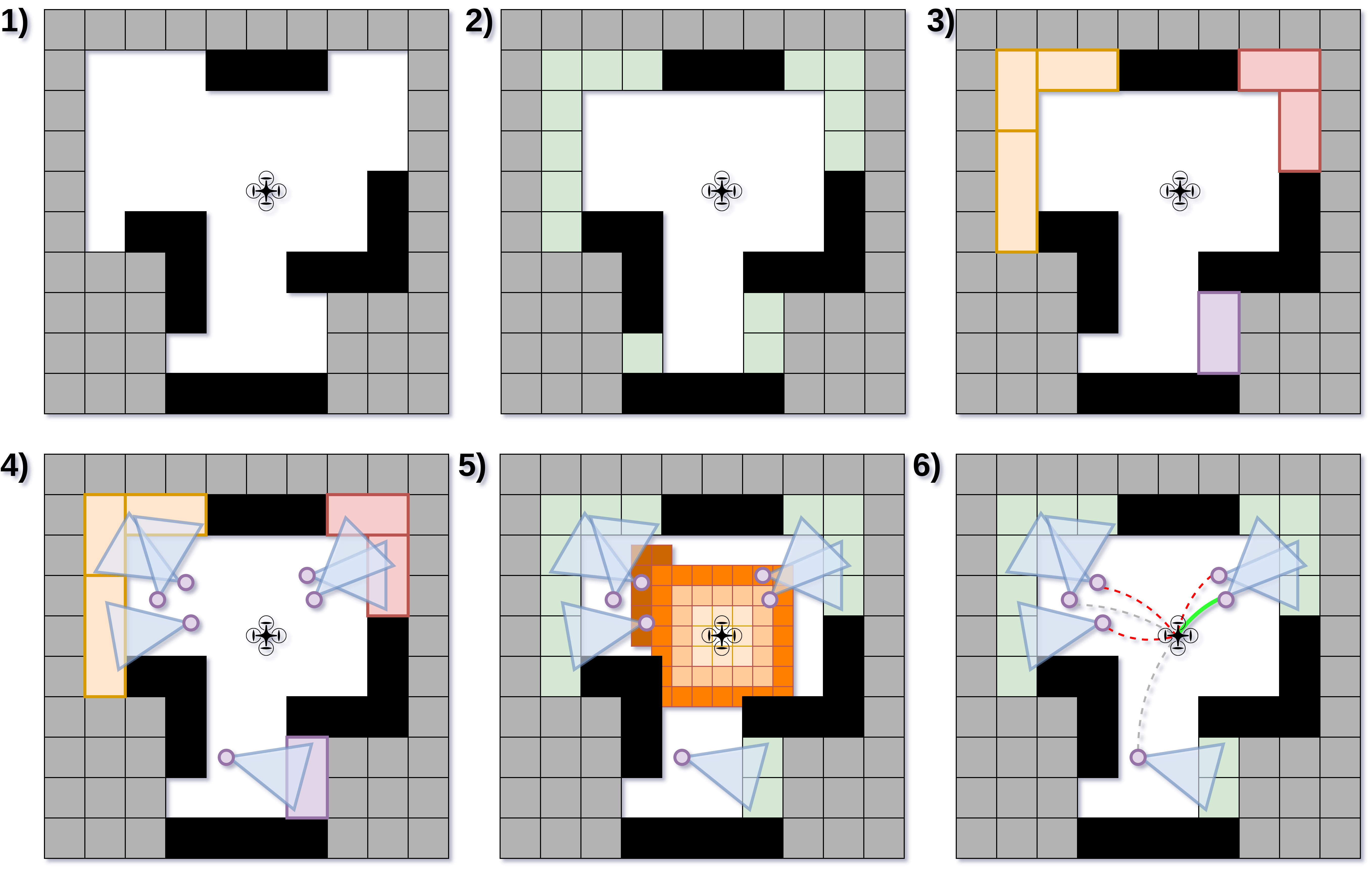}
    \caption{A 2D diagram of the frontier goal pose sampling and planning approach on an occupancy grid.  1) Planning starts with dual occupancy-grid and ESDF maps.  2) Label frontier voxels as free cells adjacent to unseen.  3) Cluster, threshold, and greedily group frontier voxels.  4) Randomly sample goal poses that view frontiers.  5) Calculate cost to a subset of goal poses until the best is found using fast marching method.  6) Follow wave front arrival time gradient to generate paths to goal poses and select the best path.} \label{fig_plan_6step}
\end{figure}

\subsection{Frontier}
In order to direct the robot's navigation towards exploring the environment, we consider sampling goal poses that focus the robot's sensor field of view on the map's frontier. A frontier is the interface between the seen and unseen portions of the environment or the set of seen and free voxels neighboring at least one unseen voxel.  More formally, the frontier is defined in (\ref{eq_frontier_definition2}) below
\begin{equation}
\label{eq_frontier_definition2}
    \mathcal{F} = \{\bm{c} : \bm{c} \in \mathcal{S}_f \text{ and } \sum_{\bm{n} \in N(\bm{c})} \begin{cases}
        1 \text{ if } \bm{n} \in \mathcal{U} \\
        0 \text{ otherwise}
    \end{cases}
    > 0\}
\end{equation}
where $\mathcal{S}_f$ is the set of seen and free voxels, $\mathcal{U}$ is the set of unseen voxels, and $N(\bm{c})$ is the 26-neighboring voxels to the voxel $\bm{c}$.

In 2D and simple 3D environments, these voxels would focus the robot's exploration quite well. However, in more complex and unstructured 3D environments, frontier voxels often show up in small orphaned groups due to sensor noise and occlusions. Focusing the robot's attention on these voxels would waste valuable flight time on exploring very small portions of the missing map. In order to rectify this issue, we consider a contiguous cluster-based filtering method to pre-select larger regions of interest for the robot to explore.

\subsubsection{Contiguous Clustering}

Frontier voxels are considered contiguous if the distance between their centers is less than or equal to $1.75s_c$. This ensures that the diagonally adjacent voxels are considered contiguous. The contiguous clustering procedure separates frontier voxels into connected clusters through a segmentation algorithm. Once the frontier voxels are grouped into their exclusive cluster sets $Q$, those above a size threshold $N_c$ are kept in the frontier subset $\mathcal{F}_{cl}$ for goal pose sampling.  Clusters are obtained using the open-source Point Cloud Library (PCL) implementation of Euclidean Cluster Extraction \cite{Rusu_ICRA2011_PCL}.

\subsection{Pose Sampling}

\begin{algorithm}[tbh]
\begin{algorithmic}[1]
  \STATE $\mathcal{P} \gets \{\}$, Groups $\gets \{\}$ \\
  \WHILE{$\{\mathcal{F}_{cl}$ \textbackslash Groups$\} \neq \{\}$}
    \STATE $\bm{c}_g \gets$ Sample$(\{\mathcal{F}$ \textbackslash Groups$\})$ \\
    \STATE $G_{\text{new}} \gets \{\}$ \\
    \FOR{$\bm{c} \in$ GetFrontierNeighbors$(\bm{c}_g, r_g)$}
      \IF{$\bm{c} \notin$ Groups}
        \STATE $G_{\text{new}} \gets [G_{\text{new}}; \bm{c}]$ \\
      \ENDIF
    \ENDFOR
    \STATE Groups $\gets [\text{Groups}; G_{\text{new}}]$ \\
  \ENDWHILE
  \FOR{$G \in$ Groups}
    \WHILE{True}
      \STATE $\bm{p}_{\text{sample}} \gets$ SamplePose$(\bar{\bm{c}}_g, [r_{min}, r_{max}], \text{FoV})$
      \IF {CheckAdmissible$(\bm{p}_{\text{sample}})$}
        \IF{!CheckOcclusion$(\bm{p}_{\text{sample}},\bar{\bm{c}}_g)$}
          \STATE $\mathcal{P} \gets [\mathcal{P}; \bm{p}_{\text{sample}}]$ \\
          \STATE \textbf{Break} \\
        \ENDIF
      \ENDIF{}
    \ENDWHILE
  \ENDFOR
\end{algorithmic}
\caption{SampleGoalPoses$(\mathcal{F}_{cl}, r_g)$}
\label{algo_ViewPoseSampling}
\end{algorithm}

The global planner samples goal poses relative to the clustered frontier using Algorithm \ref{algo_ViewPoseSampling}.  The remaining set of clustered frontier, $\mathcal{F}_{cl}$, are grouped using a greedy radius-based procedure.  Then, one goal pose is sampled .  Goal poses are sampled relative to each group's centroid voxel, $\bar{\bm{c}}_G$, using a random uniform distribution in polar coordinates, given in the equation below, where $\alpha_{\text{FoV}}$ is half the robot's vertical field of view.  A goal pose is admissible if it is far enough from an occupied voxel and its line of sight to the group centroid voxel is unoccluded.  This technique was adapted from an underwater surface inspection algorithm \cite{englot2013three}. 

\begin{equation}
\begin{aligned}
  \bm{p}_{\text{sample}} &= \begin{bmatrix}
  \bar{\bm{c}}_g \\
  -\phi
  \end{bmatrix} + \rho \begin{bmatrix}
  \sin{\alpha}\cos{\phi} \\
  \sin{\alpha}\sin{\phi} \\
  \cos{\alpha} \\
  0
  \end{bmatrix} \\
  \rho \sim \mathcal{U}(r_{\text{min}}, r_{\text{max}}), &\quad \alpha \sim \mathcal{U}(-\alpha_{\text{FoV}},\alpha_{\text{FoV}}), \quad \phi \sim \mathcal{U}(0,2\pi)
\end{aligned}
\label{eq_polar_sample}
\end{equation}

Each sampled goal pose, $\bm{p}$, is given corresponding information gain, $G(\bm{p})$.  The gain function used in this work is the number of frontier voxels in the sensor field of view at the sampled goal pose.  This encourages planning to poses that explore a large amount of unseen space.

\subsection{Fast Marching Method}
The arrival time $T$ of a wave front starting at the robot's current voxel location $c(\bm{r}$), to each voxel in the set of seen and free voxels, $\bm{c} \in \mathcal{S}_f$, is calculated using a fast approximation of the solution to the Eikonal Equation given below \cite{hassouna2007multistencils} \cite{kroon2020MSFM}.
\begin{equation}
    |\nabla T(\bm{c})|S(\bm{c}) = 1
\end{equation}
where $S(\bm{c})$ is the wave propagation speed at voxel, $\bm{c}$ given in the equation below
\begin{equation}
    S(\bm{c}) = \frac{1}{2}(\text{tanh}(D(\bm{c}) - e) + 1)
\end{equation}
This speed function penalizes paths that travel close to obstacles in order to avoid collisions and maximize the sensor field of view along the path.  This wave front propagation is run until the goal pose with the greatest utility, $U(\bm{p}) = \frac{G(\bm{p})}{T(c(\bm{p}))}$, is found.  Given that the current voxel arrival time, $T(\bm{c}_{\text{FMM}})$, is increasing with fast marching algorithm run time, there is an upper bound on the maximum utility of any goal pose in the subset of goal poses with no evaluated wave front arrival time, $\mathcal{P}_{ne}$.  Therefore, the fast marching calculation can prematurely terminate once it satisfies the following stopping criteria:
\begin{equation}
     \max_{\textbf{p} \in \{ \mathcal{P} \setminus \mathcal{P}_{ne} \}}{U(\bm{p})} \geq \frac{\max_{\bm{p} \in \mathcal{P}_{ne}} G(\bm{p})}{ T(\bm{c}_{\text{FMM}}))}
\end{equation}
This guarantees that the cost-to-go to the best goal pose has already been found without performing any extraneous computation.  After selecting the goal pose with the highest utility, the 3D Sobel operator is used to follow the voxel gradient path of the wave front arrival time, $T(\bm{c})$, from the selected goal pose back to the robot's current position.  The robot then uses this path to generate control commands.

\begin{figure}
		\includegraphics[width=.23\textwidth]{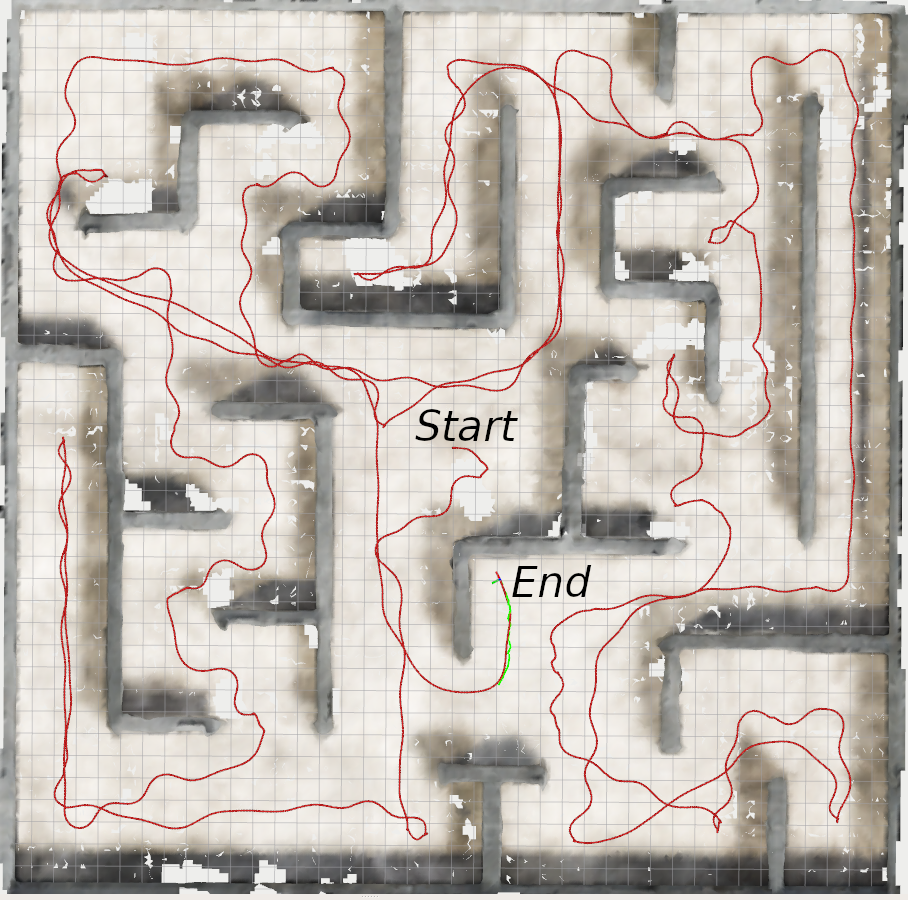}
		\includegraphics[width=.25\textwidth]{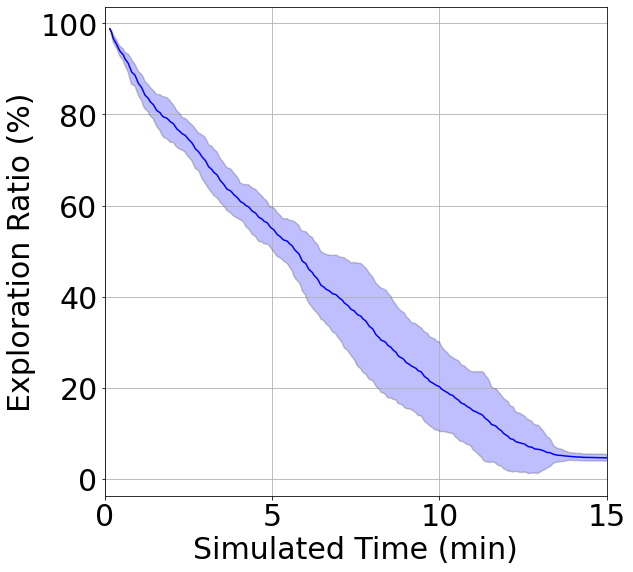}
		\caption{Exploration performance of the global planning algorithm in a simulated 40m x 40m x 3.0m Unreal Engine 3D maze environment. The UAV was constrained to a max flight speed of $1$ m/s and turn rate of $\pi/2$ rad/s and had a sensor FoV of $90^{\circ}$ x  $73.7^{\circ}$ and a range of $5$m. The left figure shows the path traveled by the simulated UAV.  The UAV tends to center itself in the maze corridors because of the ESDF speed map which leads to efficient flight paths and large sensor fields of view.  The right figure shows the mean exploration volume ratio vs time and $2\sigma$ bounds for $10$ simulation runs.}
		\label{fig:Planning_maze_simulation_volume_vs_time}
\end{figure}

Paths produced by following the wave arrival time map gradient to each frontier goal pose are not only obstacle-aware, but they also tend to produce en-route poses that reduce depth sensor occlusion over the length of the path.  This results in a greater overall exploration rate because the sensor field of view is implicitly maximized by penalizing obstacle proximity when calculating goal pose costs.

The global planning algorithm is validated in a simulated environment same as that of \cite{schmid2020efficient} which utilizes a UAV with a single RGBD sensor. The exploration ratio across $10$ simulation runs is shown in Fig. \ref{fig:Planning_maze_simulation_volume_vs_time} along with the path taken by the UAV for one of the simulation runs. The planner was able to explore the environment in less than $14$ minutes during each run.

\section{Local Reactive Control}
\label{sec:local_planning}

The goal of the reactive control problem is to navigate the UAV along the global path safely while respecting the vehicle kinematics. This planning layer ensures that the robot is \textit{repelled} away from the obstacles that would not otherwise be avoided due to imperfect localization, mapping and path following. To this end, we exploit the APF technique using direct sensor information from the forward, upward and downward-facing depth cameras onboard the UAV. The repulsive potential acting on the robot is the sum of the repulsive potentials from the individual cameras transformed to the robot's body frame $\mathcal{F}_\mathcal{B}$. \par

We assume that any 3D point $\bm{p} = (p_x,p_y,p_z)$ in a depth camera's frame can be projected to its image coordinates $(s_x,s_y)$ using the pinhole camera model as
\begin{align}
    \label{eq:pinhole_model}
    s_x &= f_x p_x / p_z + c_x, \nonumber \\
    s_y &= f_y p_y / p_z + c_y,
\end{align}
where $(f_x,f_y)$ defines the focal length of the camera lens and $(c_x,c_y)$ is the location of the optical center of the camera in the image coordinates. It is important to note that the number of depth image points, representing a scene object, increases as the inverse square of the depth of the object for a pinhole camera. For instance, a rectangular obstacle of length $p_x$ and height $p_y$ located at depth $p_z$ parallel to the camera's optical frame has the cross-sectional area of $p_xp_y$. The number of pixels it occupies inside the image can be computed as the occupied image area as $\frac{p_xp_yf_xf_y}{p_z^2}$. However, we would like the repulsive potential function to be dependent on the object's distance and not the number of points it occupies inside the image. Therefore, it does not seem suitable to sum the conventional APF repulsive potential gradient vectors generated by all the image points.  \par

Given a pointcloud of size $N$ from a camera, we calculate the depth image-based repulsive potential gradient vector using a slightly modified form as
\begin{align}
    \label{eq:repulsive_vector}
    \bm{u}_{\rep} = \sum \limits_{i=0}^N  k_{\rep} \bigg(\frac{1}{p_z^{\max}} - \frac{1}{p_z^i}\bigg) \hat{\bm{p}}_i,
\end{align}
where $k_{\rep}$ is a positive constant, $p_z^i$ is the depth of the $i$th pixel and $p_z^{\max}$ refers to the maximum sensing horizon. By using (\ref{eq:repulsive_vector}), all the points inside a depth image are implicitly treated as a small obstacle which is exerting a single repulsive potential regardless of the number of depth image points. Assuming the small rectangular obstacle and a camera of infinite resolution, the depth image-based repulsive potential gradient vector (\ref{eq:repulsive_vector}) reduces to
\begin{align}
    \label{eq:repulsive_vec_approx}
    \bm{u}_{rep} &\approx \int_0^{p_y} \int_0^{p_x} k_{rep} \bigg(\frac{1}{p_z^{\max}} - \frac{1}{p_z}\bigg) \hat{\bm{p}} \: dxdy, \nonumber \\
    &= k_{rep} \bigg(\frac{1}{p_z^{\max}} - \frac{1}{p_z}\bigg) \frac{p_xp_yf_xf_y}{p_z^2} \hat{\bm{p}},
\end{align}

taking integral of which with respect to $p_z$ yields the repulsive potential function in its conventional form \cite{khatib1986real} as if it were generated using a single obstacle. \par

The global planner generates a lookahead point that continually updates to be a constant distance away from the robot. The attractive potential gradient vector is applied to the robot in the direction of the lookahead point $\hat{\bm{x}}_{\text{lookahead}}$ in $\mathcal{F}_\mathcal{B}$ at all times. Consequently, the resultant acting on the UAV is given as,

\begin{align}
    \label{eq:resultant_vec}
    \bm{u}_{\text{res}} = (1-|\bm{u}_{\text{rep}}^\mathcal{B}|/|\bm{u}_{\text{rep}}^{\max}|) \hat{\bm{x}}_{\text{lookahead}} + \bm{u}_{\text{\text{rep}}}^\mathcal{B}/|\bm{u}_{\text{\text{rep}}}^{\max}|,
\end{align}
where $\bm{u}_{\text{rep}}^\mathcal{B}$ represents the sum of the individual repulsive potential gradient vectors transformed to $\mathcal{F}_\mathcal{B}$. The term $1-|\bm{u}_{\text{rep}}^\mathcal{B}|/|\bm{u}_{\text{rep}}^{\max}|$ regulates the speed of the UAV depending upon the level of clutter around it. The user-defined $|\bm{u}_{\text{rep}}^{\max}|$ threshold can be set based on the desired aggressiveness of the robot in reaction to the nearby obstacles. The velocity commands for the UAV are finally generated to follow the resultant 3D vector as

\begin{align}
    \label{eq:uav_vel_cmds}
    v_x &= v_x^{\max} u_{\text{res}}^x, \nonumber \\
    v_z &= v_z^{\max} u_{\text{res}}^z, \nonumber \\
    v_{\theta} &= \frac{1}{\pi} v_{\theta}^{\max} \text{atan2} (u_{\text{res}}^y,u_{\text{res}}^x),
\end{align}
where $v_x$, $v_z$ and $v_{\theta}$ are the forward, vertical and steering velocity commands in $\mathcal{F}_\mathcal{B}$ respectively and $v_x^{\max}$, $v_z^{\max}$ and $v_{\theta}^{\max}$ define the maximum velocities for each axis.

\section{Experimental Evaluation}
\label{sec:experimental_evaluation}

\begin{figure}[h!]
		\centering
		{{\includegraphics[width=0.48\textwidth]{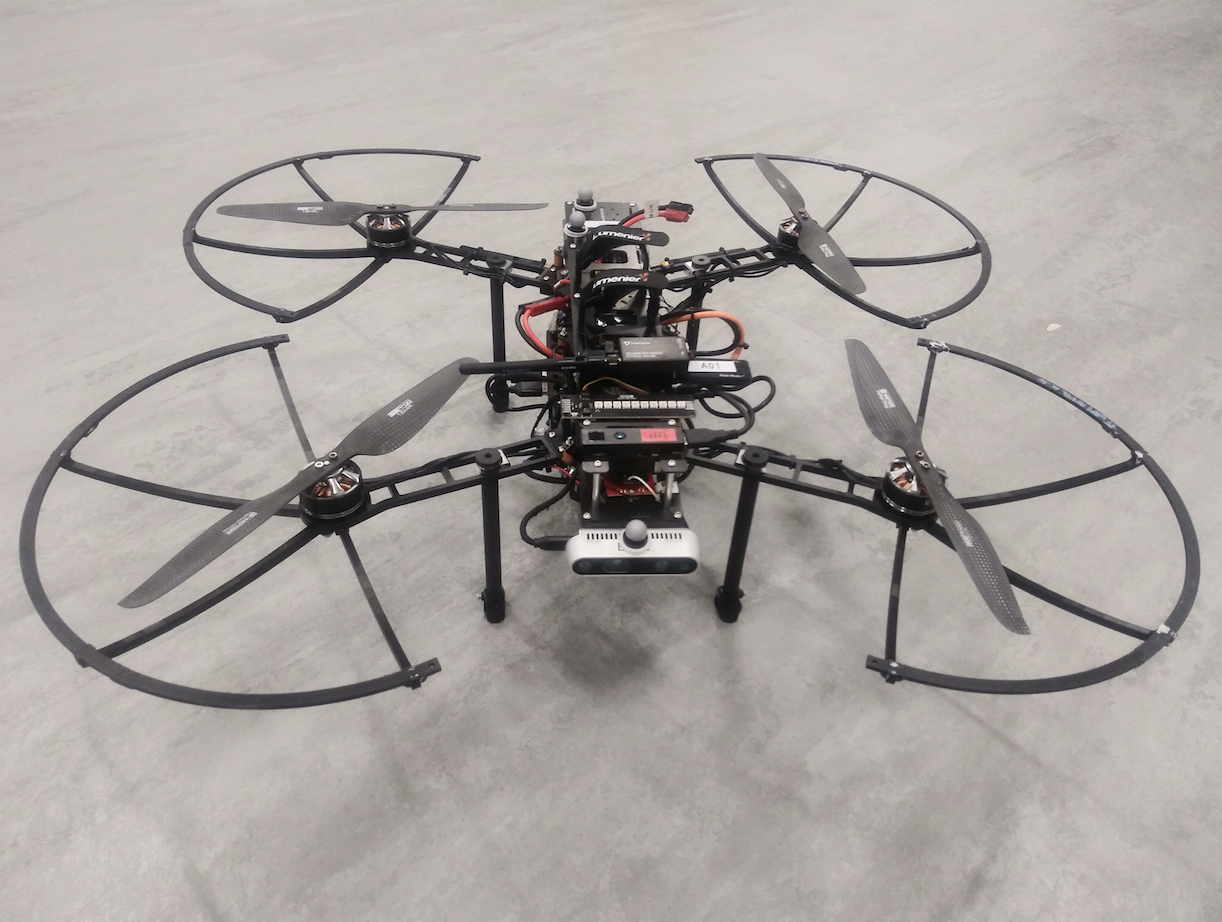}\label{fig:UAV_ASPEN_Trajectory_2}}}
		\caption{The quadrotor UAV platform used for the hardware experiments.}
		\label{fig:Photo_UAV}
\end{figure}

To evaluate the performance of the proposed exploration and control strategies, experiments were conducted on a UAV platform, as shown in Fig. \ref{fig:Photo_UAV}. The demands of real-time operation, along with size, weight, and power considerations, limit the computation available for autonomy.

\subsection{Aerial Robot Platform}

The UAV control system begins with the Pixracer R15, a PX4-based flight controller that provides attitude stabilization and velocity control. For sensing, an Ouster OS-1-64 LiDAR and Lord Microstrain 3DM-GX-15 IMU support localization and mapping, while a forward-facing Intel RealSense D435i, as well as upward-facing and downward-facing Camboard pico flexx depth cameras, are utilized for obstacle avoidance. Several open-source software packages, including Google Cartographer \cite{hess2016} for LiDAR-inertial SLAM, VoxBlox \cite{oleynikova2017voxblox} for ESDF mapping, and OctoMap \cite{hornung13octomap} for mapping visualization, provide input to the proposed global and local planners, which all operate onboard a Intel NUC 7i7DNBE computer.

\subsection{Local Control Demonstration}

Worse-case local control scenarios occur when mapping fails to capture obstacles in the environment, and frontier-based exploration plans an infeasible path that would collide with an undetected obstacle. In particular, small or thin obstacles, such as cabling and wiring, are less likely to be captured in the map. More generally, localization, mapping, and path following errors can also lead to close encounters or collisions. In scenarios such as these, it is the responsibility of the local planner to rapidly sense the proximity of obstacles, and ensure an evasive maneuver is taken. 

\begin{figure}[h!]
		\centering
		\subfloat[]{{\includegraphics[width=0.65\linewidth,angle=270]{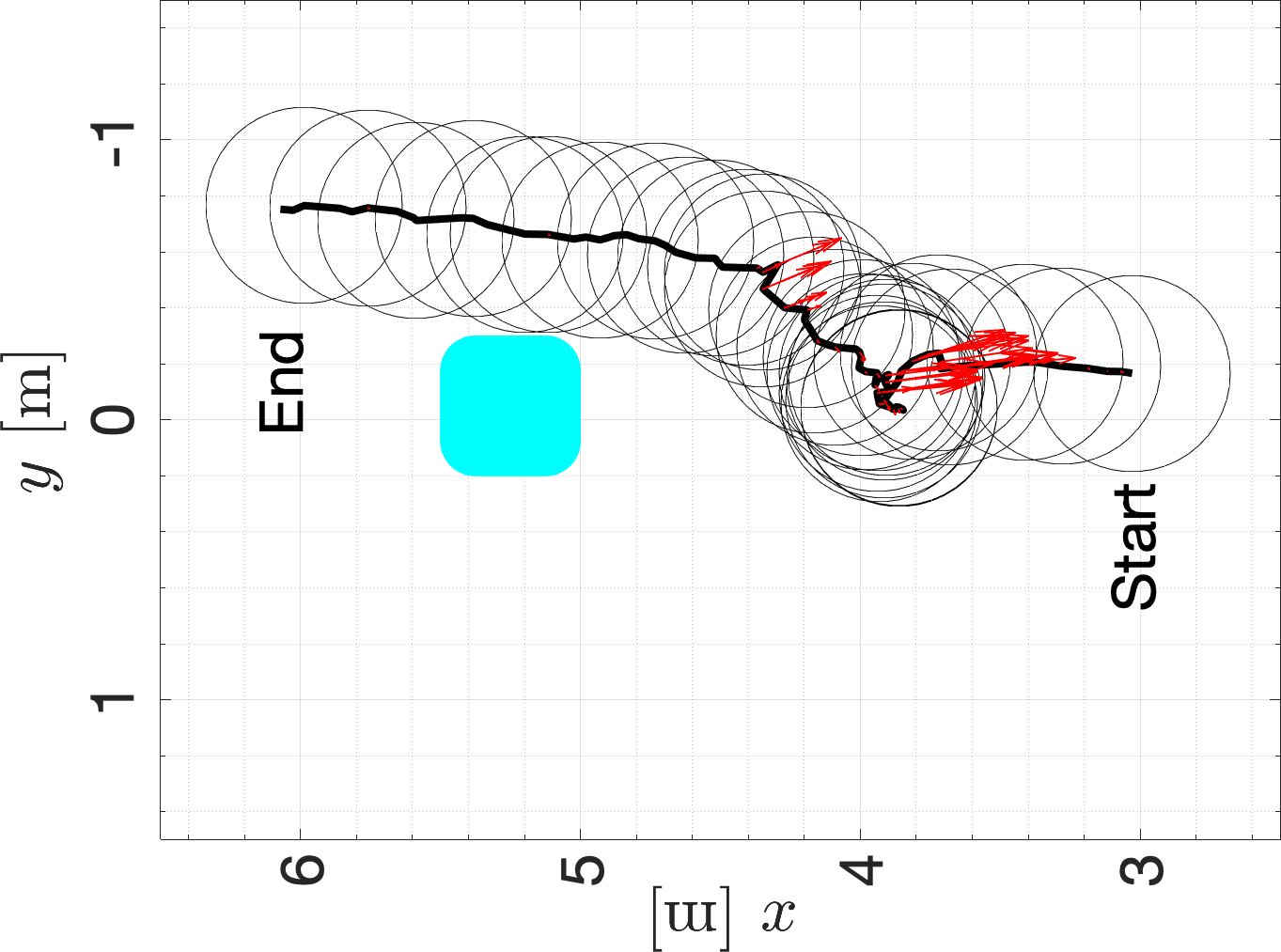}\label{fig:ASPEN_Tripod_Trajectory_2}}}
		\hspace{1pt}
		\subfloat[]{{\includegraphics[width=0.65\linewidth,angle=270]{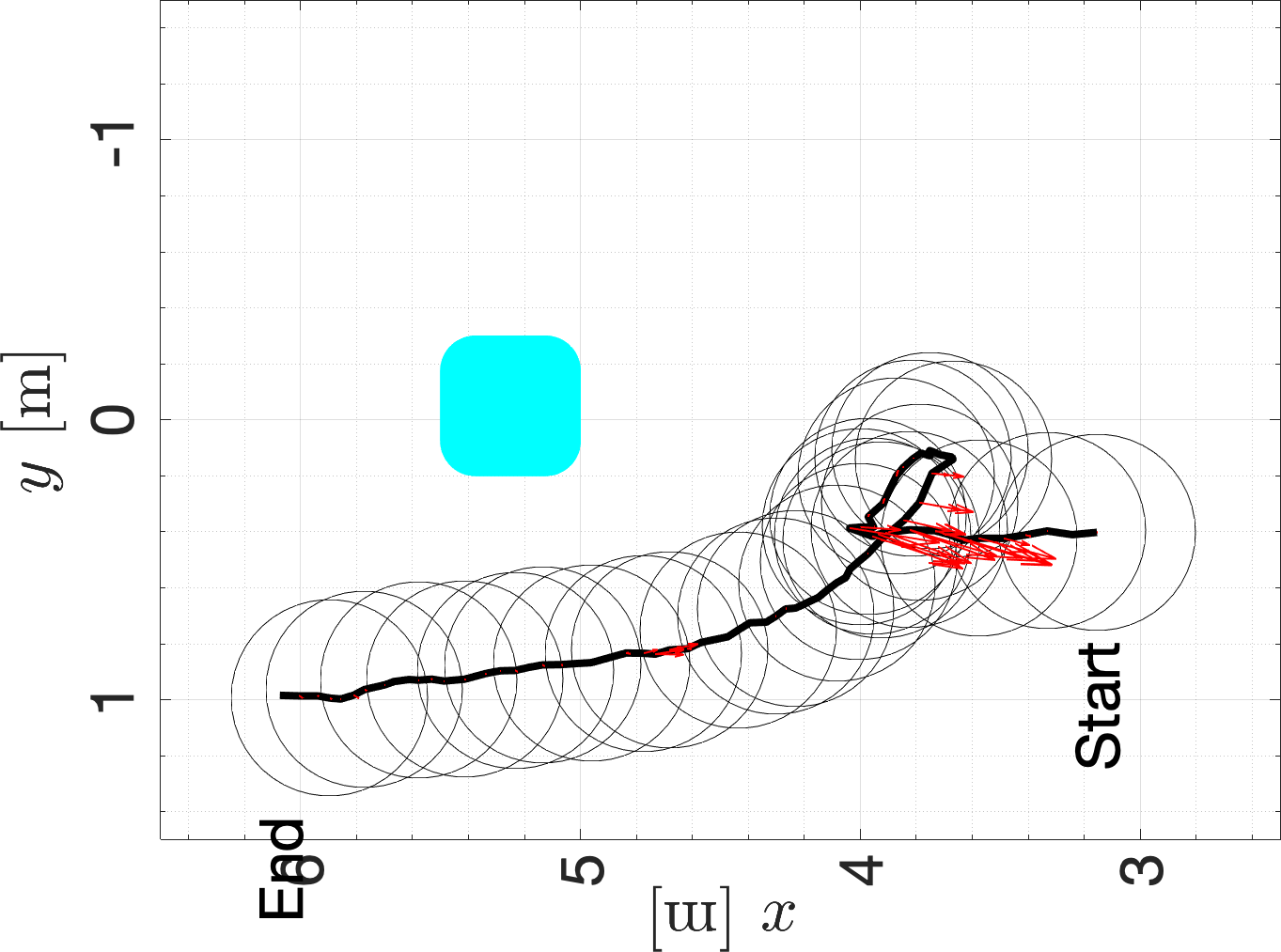}\label{fig:ASPEN_Tripod_Trajectory_6}}}
		\newline
		\caption{Results from a single obstacle local planner flight tests. The black circles represent the UAV physical bounds and the rounded cyan square represents the obstacle. The black lines show the path followed by the UAV during each of the maneuvers with red arrows representing the repulsive potential gradient vectors. The UAV starts from the respective start locations for each of the runs and maneuvers towards the goal points that are (a) 2 m and (b) 5 m behind the obstacle. The approximate obstacle location is overlaid on the plots.}
		\label{fig:uav_local_control_tripod}
\end{figure}

\begin{figure}[h!]
		\centering
		\subfloat[]{{\includegraphics[width=0.50\textwidth]{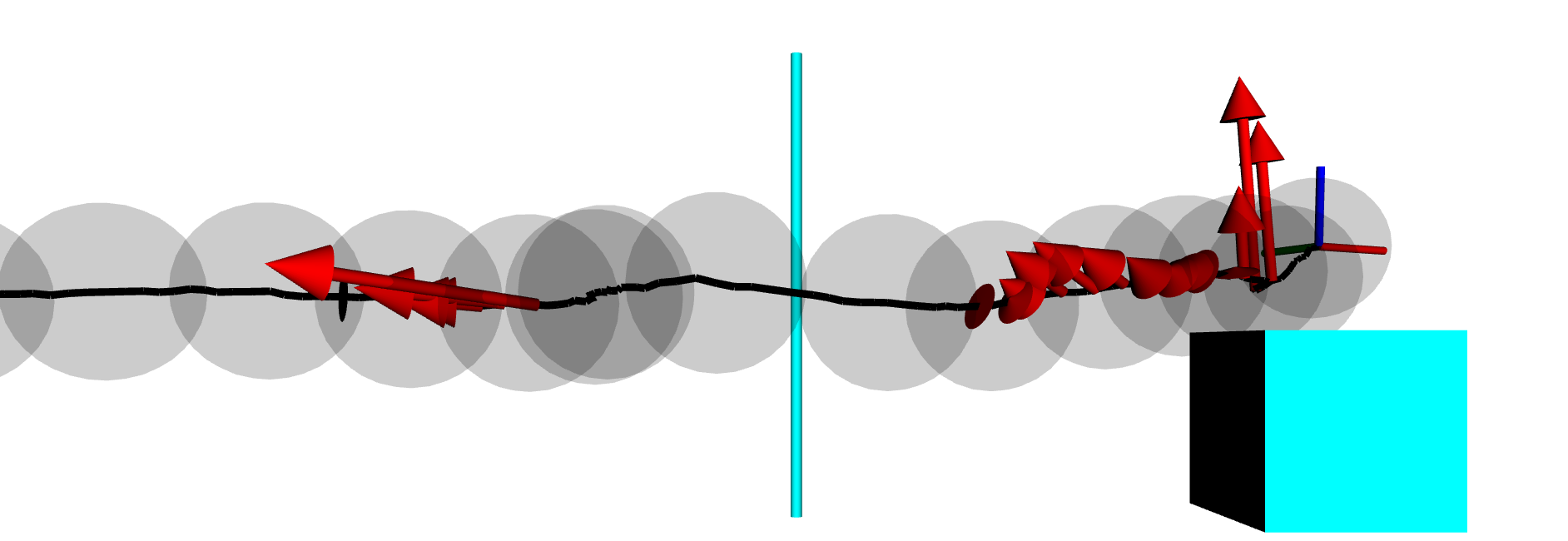}
		\label{fig:ASPEN_XZ_POLE_BOX}}}
		\newline
		\subfloat[]{{\includegraphics[width=0.50\textwidth]{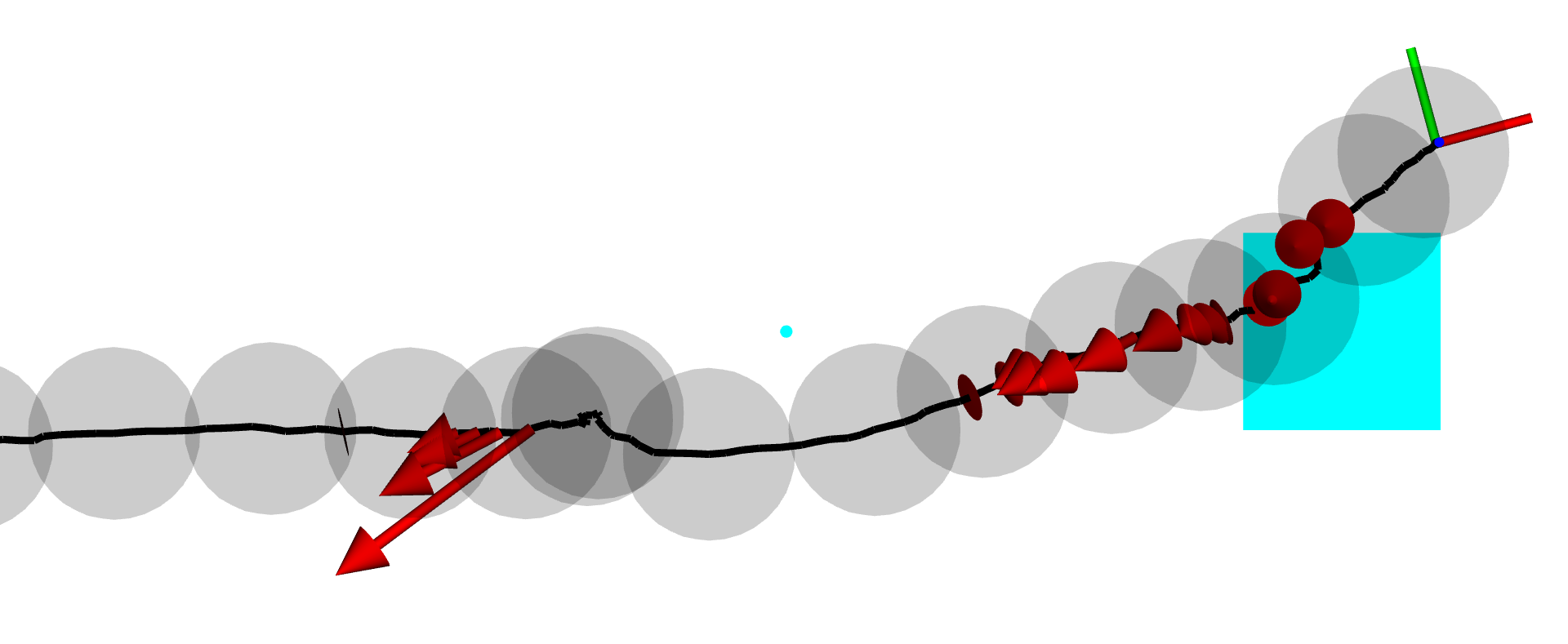}
		\label{fig:ASPEN_XY_POLE_BOX}}}
		\caption{The local planner triggers the repulsive potentials when the UAV is within close proximity of a narrow 22 mm wide pole, and a 80 cm tall, vertically-prominent box. The response is shown from a (a) side view, and (b) top-down view. The traversed trajectory, spherical vehicle bounds and the repulsive potential gradient vectors are represented as the black lines, the gray spheres, and the red arrows respectively, The approximate locations and geometries of the obstacles are shown by the solid team objects.}
		\label{fig:uav_local_control_box_pole}
\end{figure}

To evaluate the performance of the local controller, we placed obstacles of different shapes and sizes in our lab and let the UAV sense and avoid them. Manual goal points are placed behind obstacles, causing the attractive potential field to pull the UAV directly towards the obstacle. Once the agent senses the object's relative proximity, the repulsive potentials push the vehicle in the opposing direction. The resulting trajectories and repulsive potential gradient vectors from a single-obstacle scenario are presented in Fig. \ref{fig:uav_local_control_tripod}. In all evaluations, the competing attractive and repulsive potentials cause the agent to decelerate and steer away from the obstacle, before continuing towards the goal point. A second, more complex scenario is visualized in Fig. \ref{fig:uav_local_control_box_pole}, which has been designed to test the controller's ability to sense and avoid thin obstacles, as well as near-field objects underneath the vehicle.
		
\subsection{Exploration Demonstration}

This experimental scenario is set in a 100x80x6m warehouse which houses manufacturing operations. Unlike the typical structured warehouse, this environment is highly cluttered and heterogeneous. Many different activities take place on the manufacturing floor, leading to object diversity that likely ranges in the hundreds or thousands. Looking up, steel-truss ceilings support the roof, featuring exposed sprinkler, ductwork, and conduit networks, as well as light fixtures and ceiling-to-floor electrical cabling. In case of a UAV navigating in such an environment, even a single close contact of the robot with an obstacle, for instance an electric cable, can prove detrimental to the exploration mission safety. \par

\begin{figure*}[h!]
		\centering
		\subfloat[]{{\includegraphics[width=0.325\textwidth]{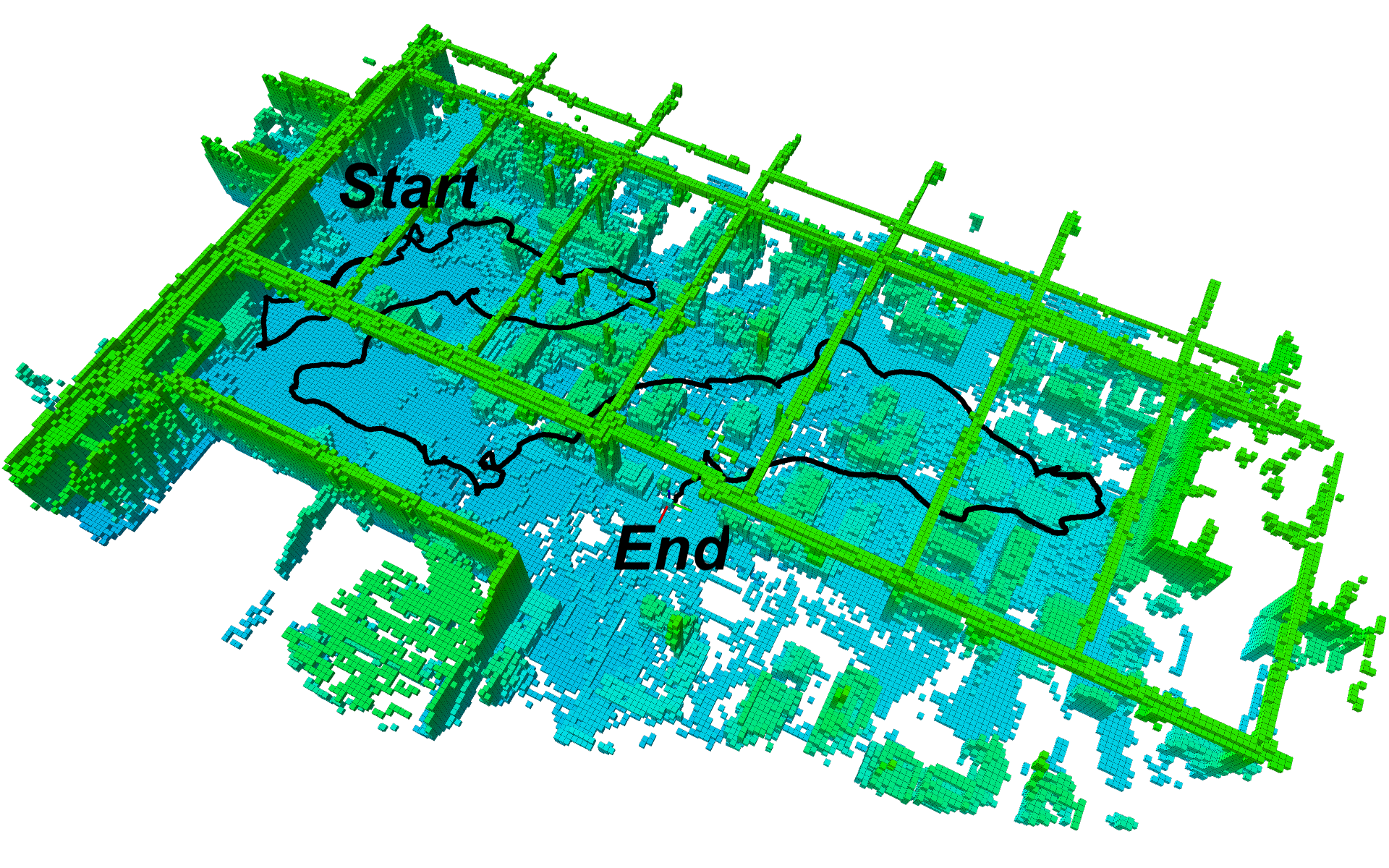}\label{fig:GEOTECH_TEST1_COMPLETE}}}
		\hspace{2pt}
		\subfloat[]{{\includegraphics[width=0.325\textwidth]{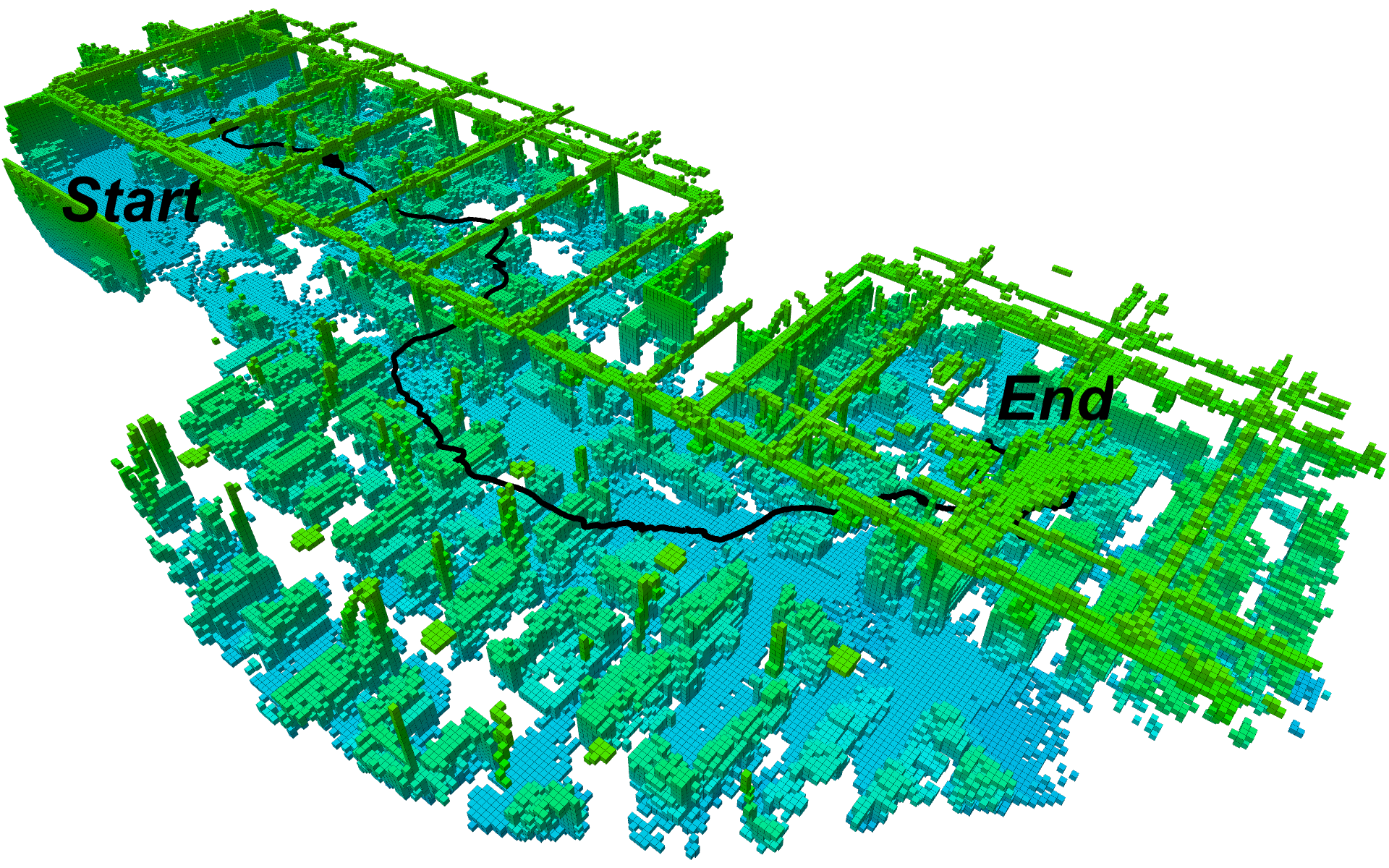}\label{fig:GEOTECH_TEST2_COMPLETE}}}
		\hspace{2pt}
		\subfloat[]{{\includegraphics[width=0.325\textwidth]{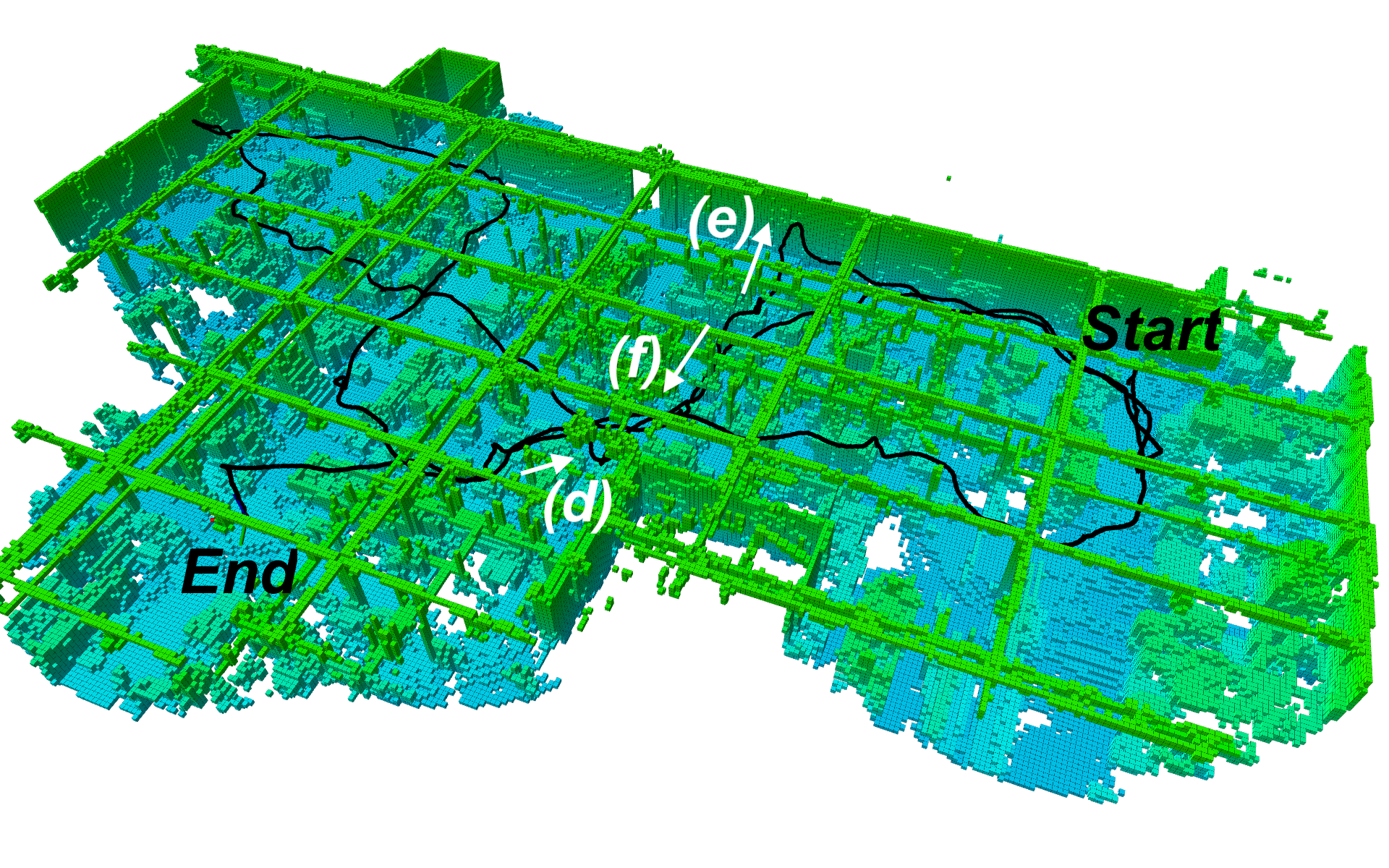}\label{fig:GEOTECH_TEST3_COMPLETE_final}}}
		\newline
		\subfloat[]{{\includegraphics[width=0.325\textwidth]{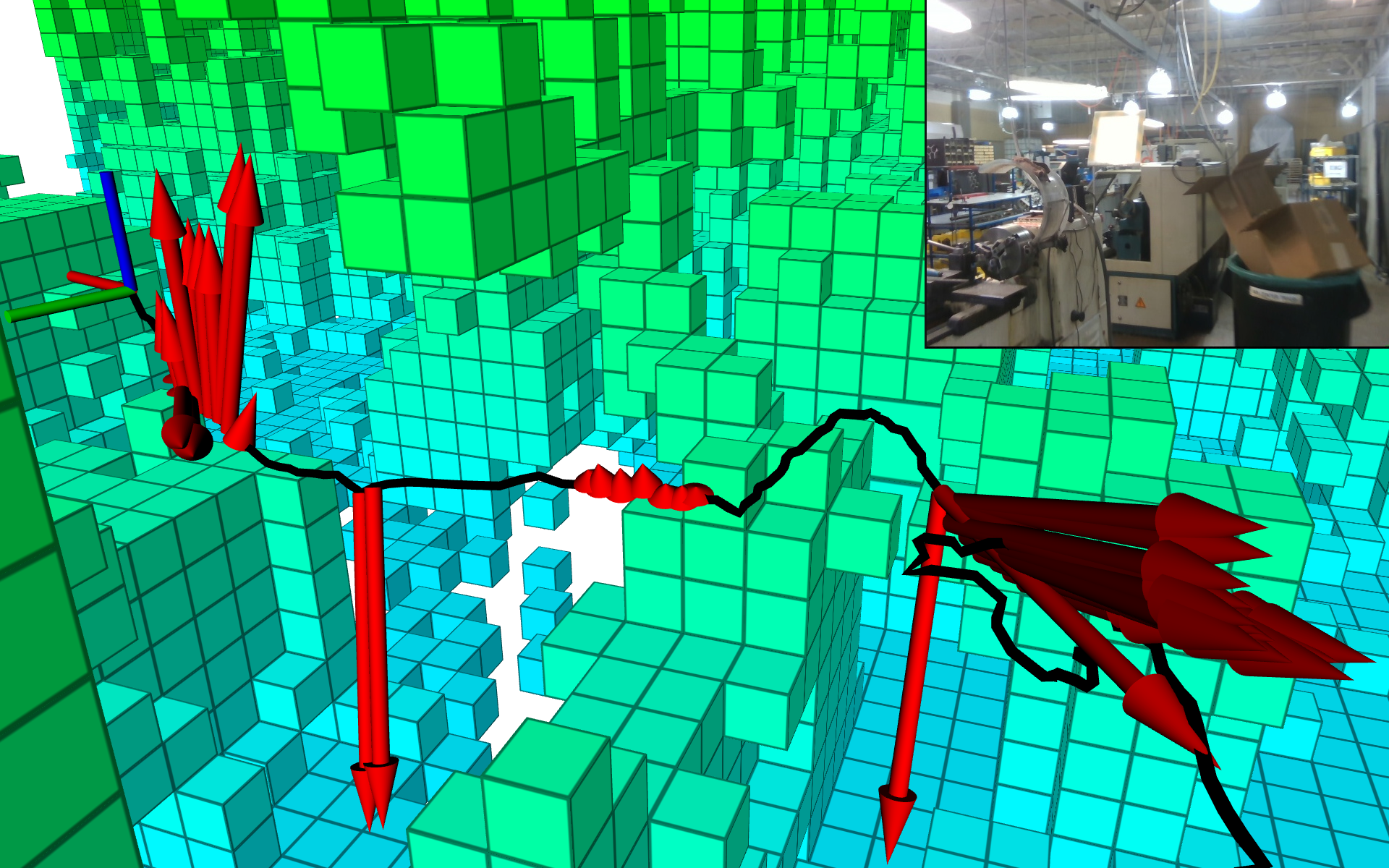}\label{fig:GEOTECH_TEST3_REPULSIVE_LATHE_OBSTACLE}}}
		\hspace{2pt}
		\subfloat[]{{\includegraphics[width=0.325\textwidth]{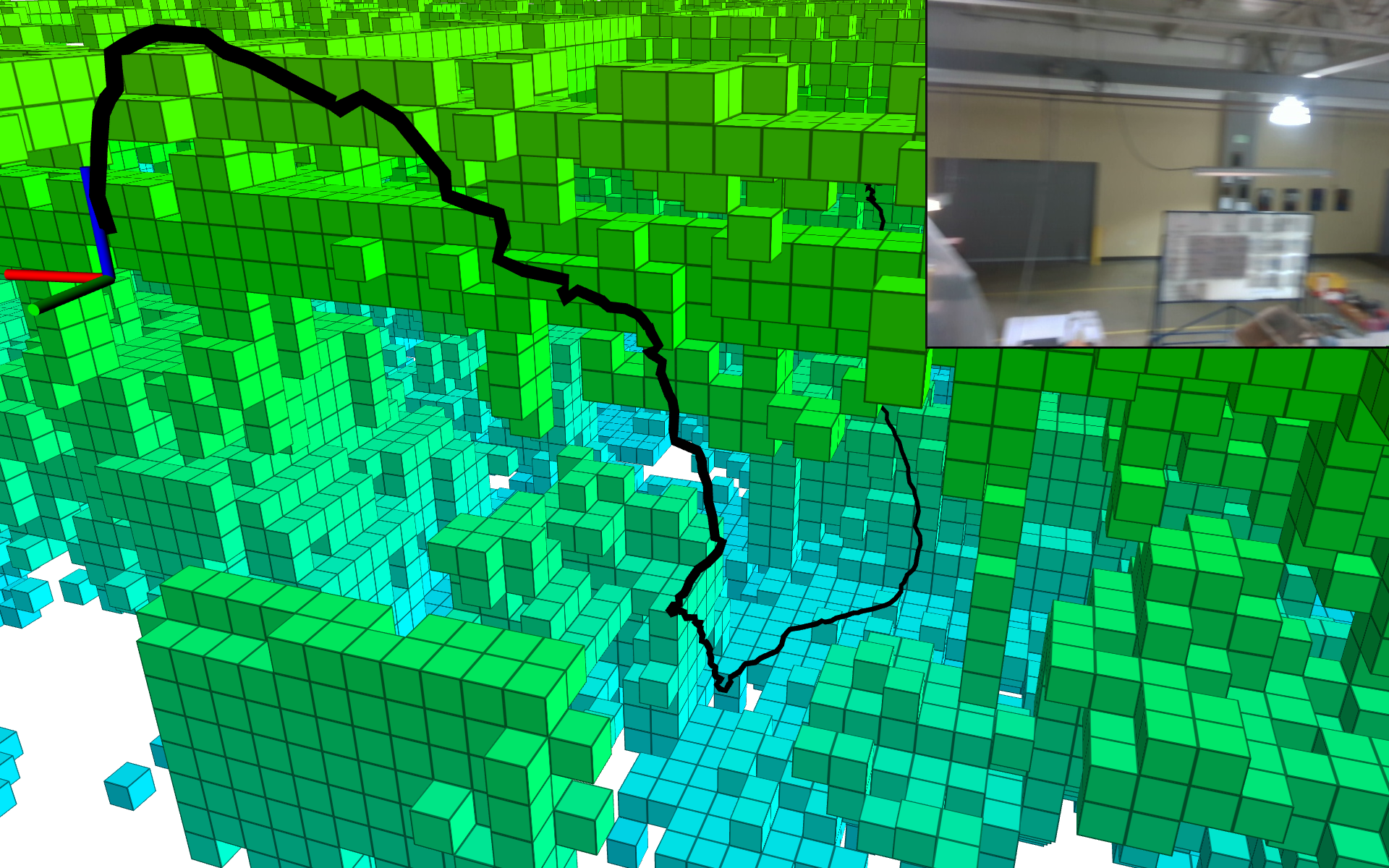}\label{fig:GEOTECH_TEST3_3D_EXPLORE_WINDOW}}}
		\hspace{2pt}
		\subfloat[]{{\includegraphics[width=0.325\textwidth]{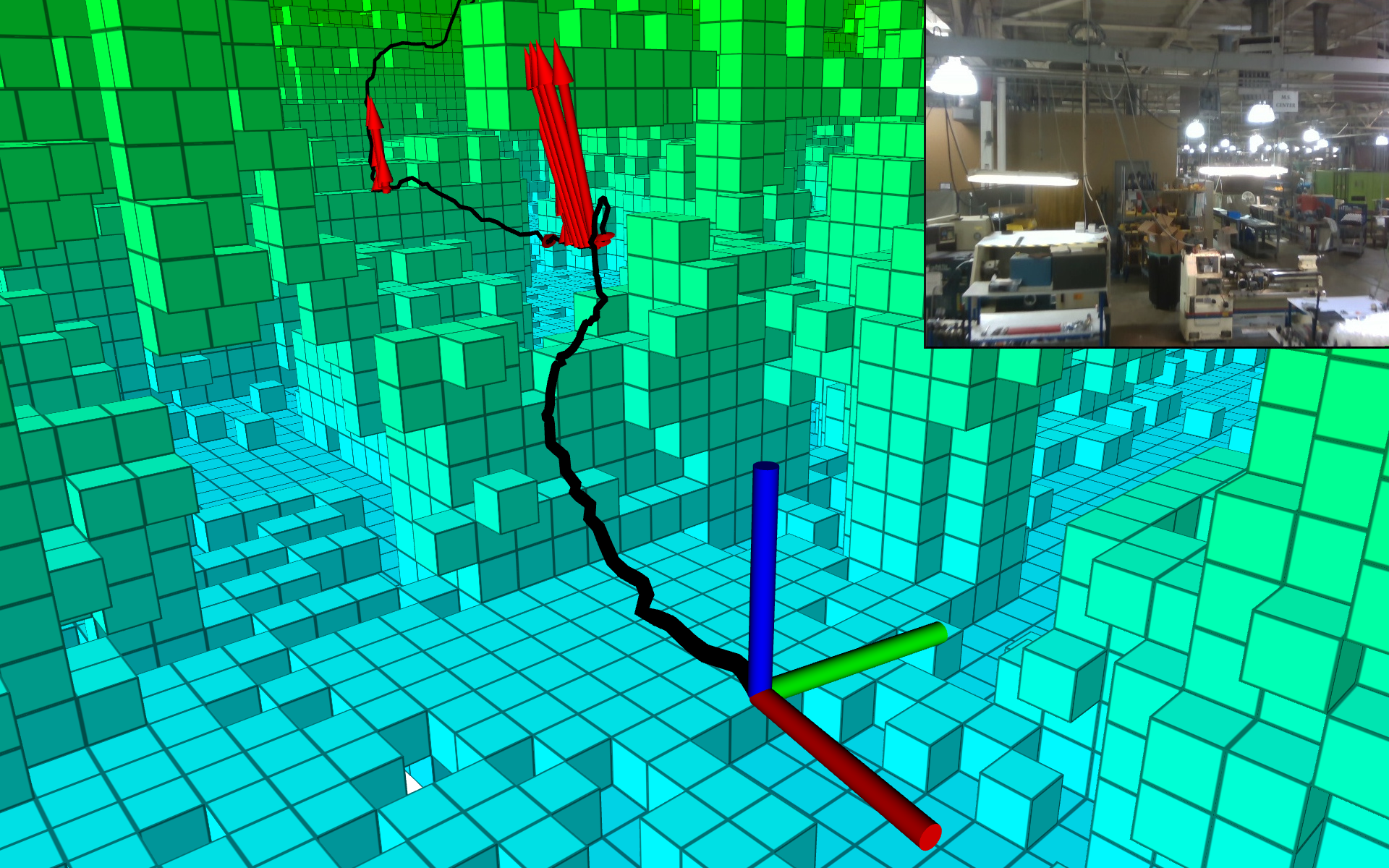}\label{fig:GEOTECH_TEST3_REPULSIVE_LATHE2_OBSTACLE}}}
		\newline
		\caption{Maps and trajectories (a, b, c) from three exploration missions in an unstructured warehouse environment, with detailed visualizations and first-person-view images during salient maneuvers (d, e, f). UAV trajectories are shown as black lines, repulsive potential gradient vectors as red arrows, and resulting positions as axes. In (c) the third field deployment, (d) the UAV maneuvered through a tight space, by flying above a lathe machining tool while remaining below low-hanging fluorescent light fixtures. A safety shield protruding up from the lathe was likely difficult to map due to its thinness, but the local planner sensed its proximity and reactively pushed the vehicle away from it. Soon after, (e) the global planner led the vehicle away from that highly-cluttered environment, by sending it through a narrow slot, and up to an open, unexplored area. The UAV returns to the machine area in (f), and again flies directly above the lathe, and is pushed up by the repulsive potential field. For explicit visualization of the occupancy grid, the OctoMaps with 0.2 m leaf size are shown. The ceiling has been removed to provide visibility of the building interior. The distances travelled by the UAV in the three deployments are (a) 152.5 m (b) 118.0 m (c) 304.08 m. }
		\label{fig:GEOTECH_TEST1_TEST2_TEST3_WITH_INSETS}
\end{figure*}

\begin{figure*}[h!]
		\centering
		\includegraphics[width=.32\textwidth]{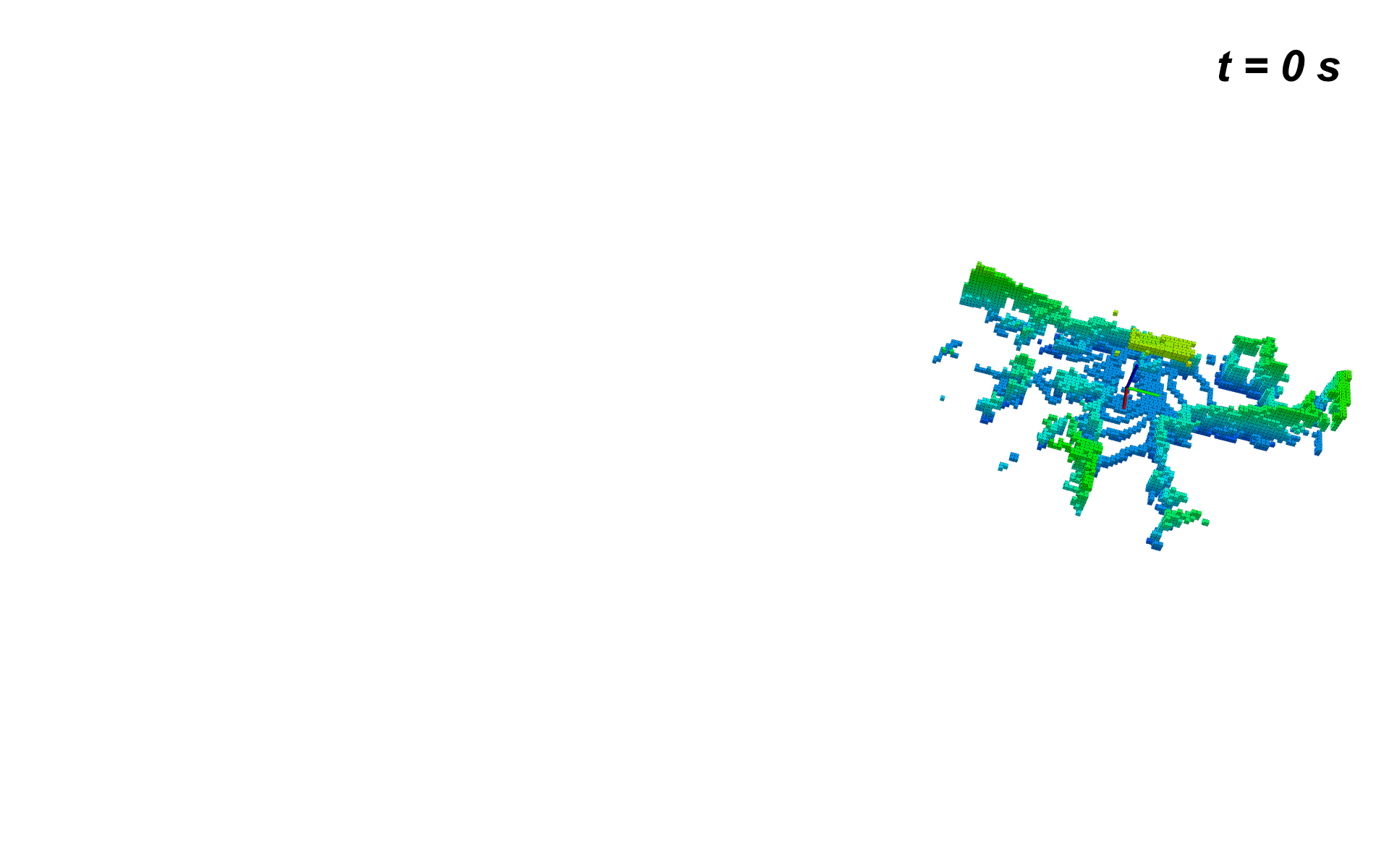}
		\includegraphics[width=.32\textwidth]{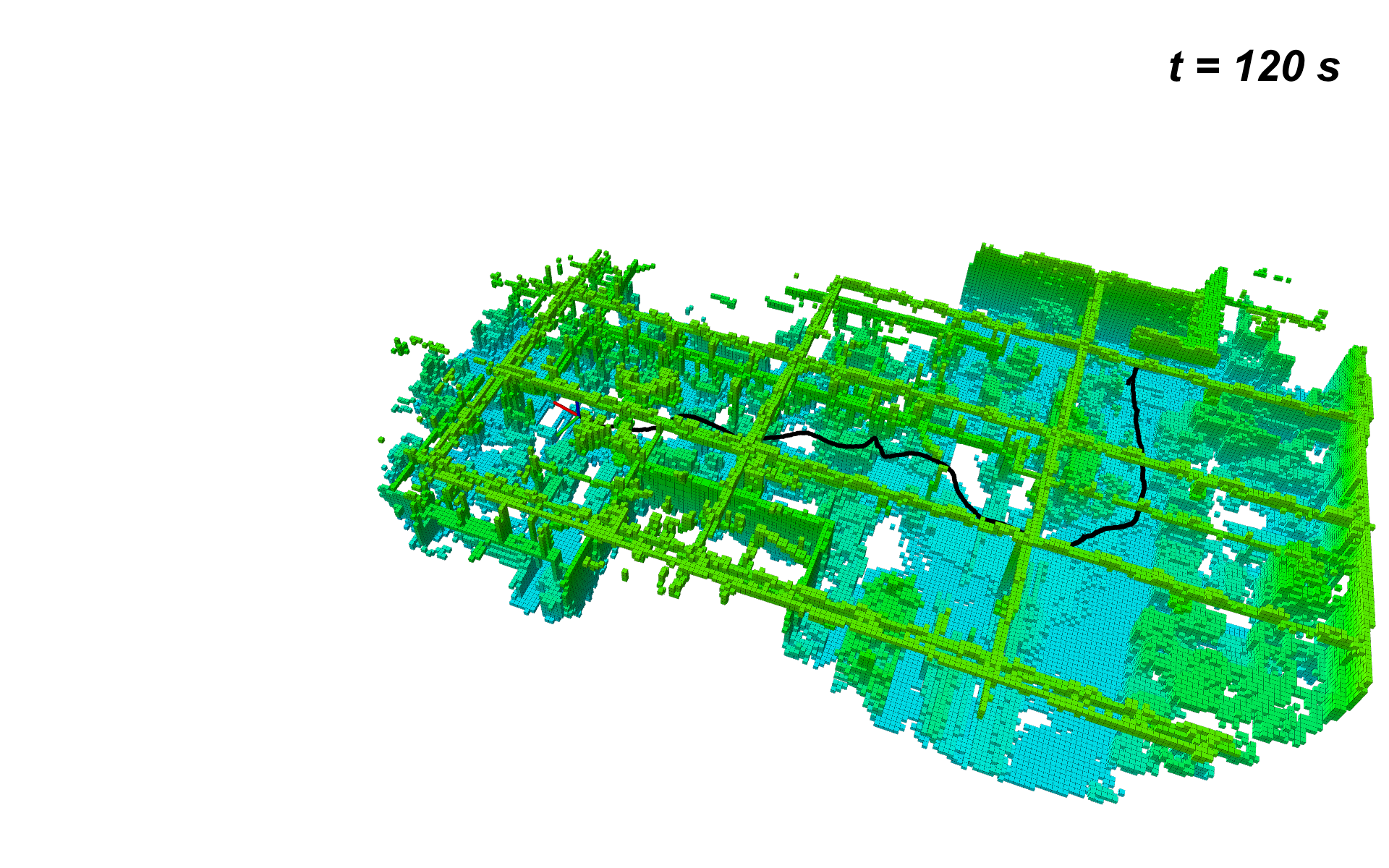}
		\includegraphics[width=.32\textwidth]{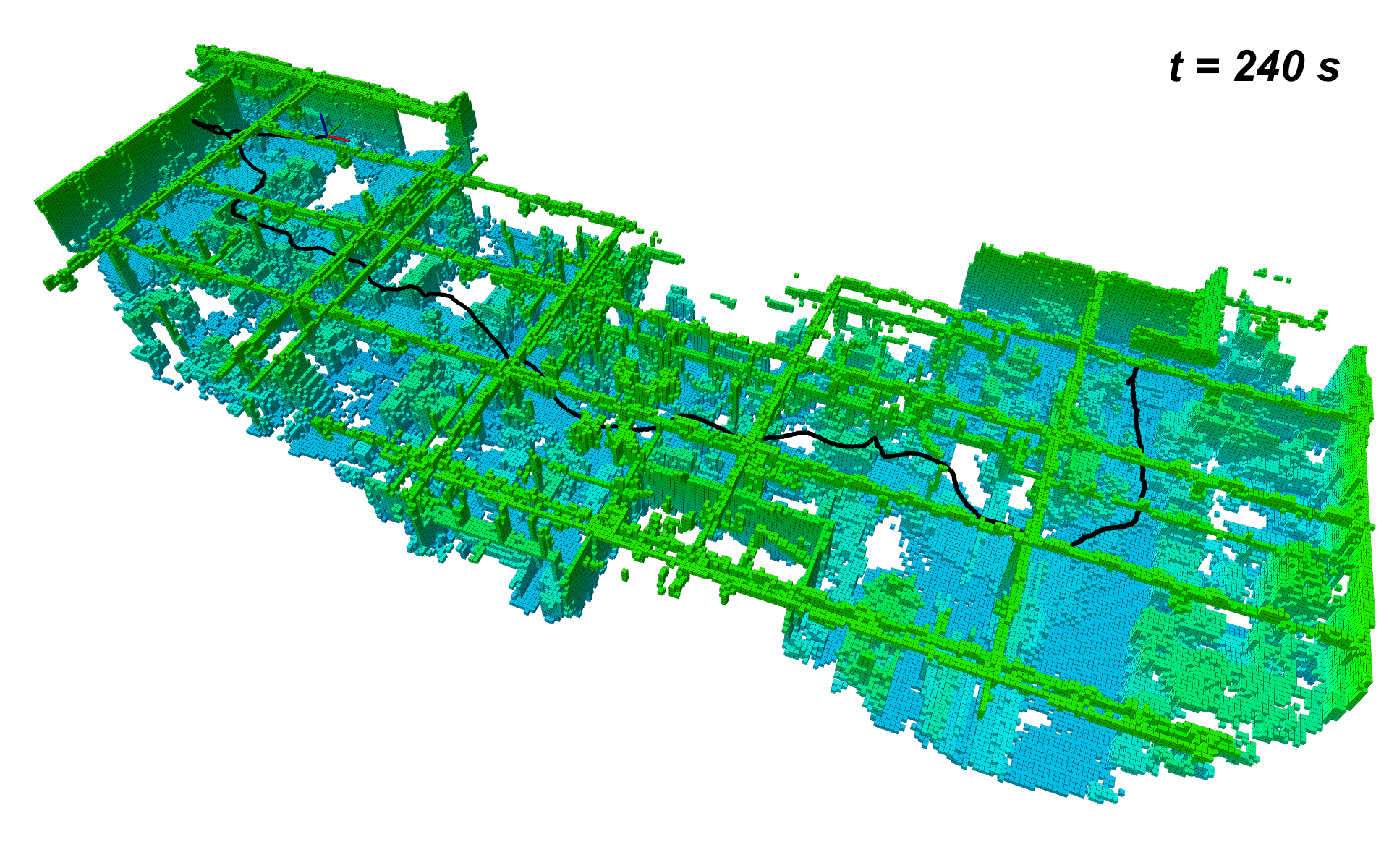}
		\newline
		\includegraphics[width=.32\textwidth]{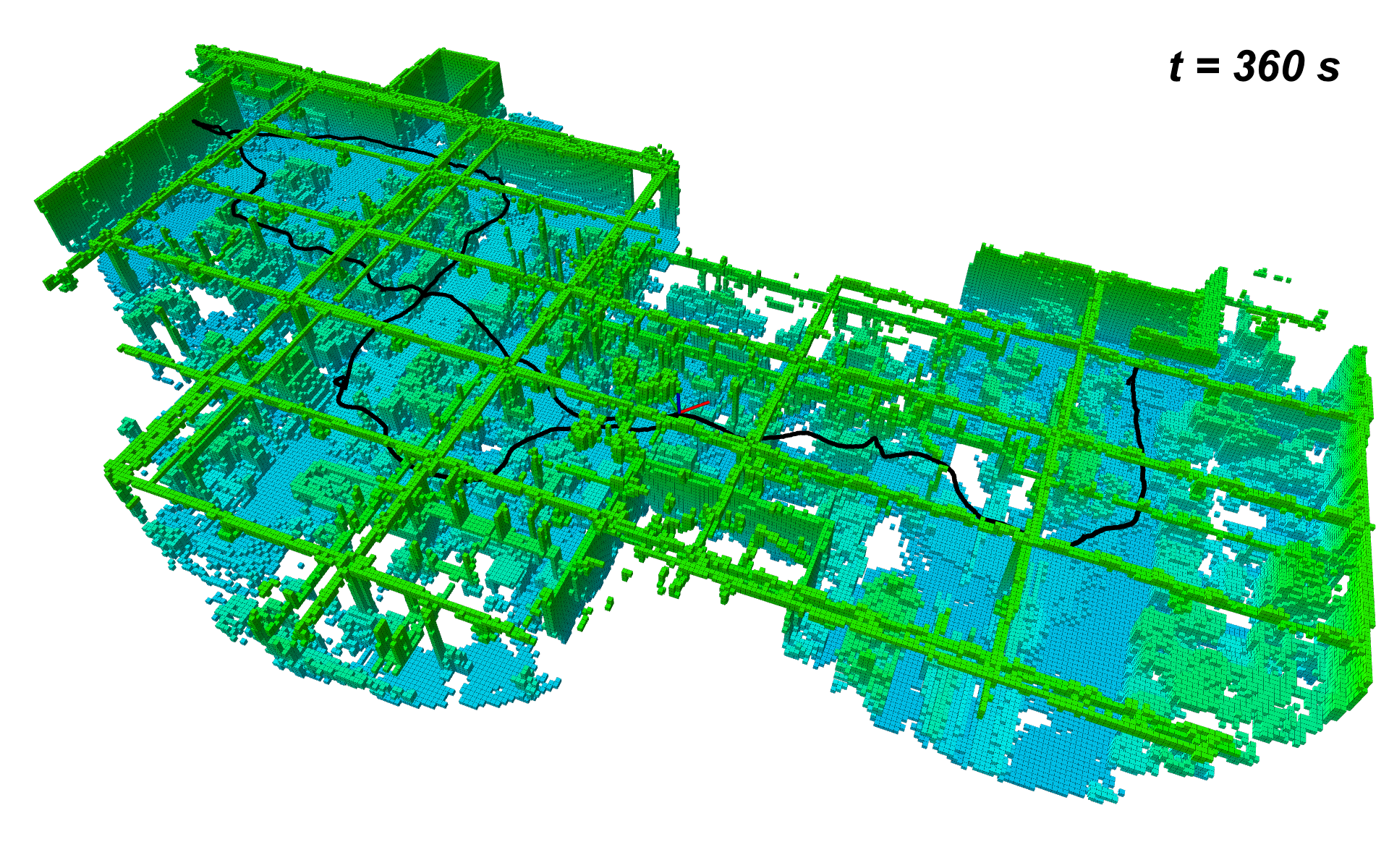}
		\includegraphics[width=.32\textwidth]{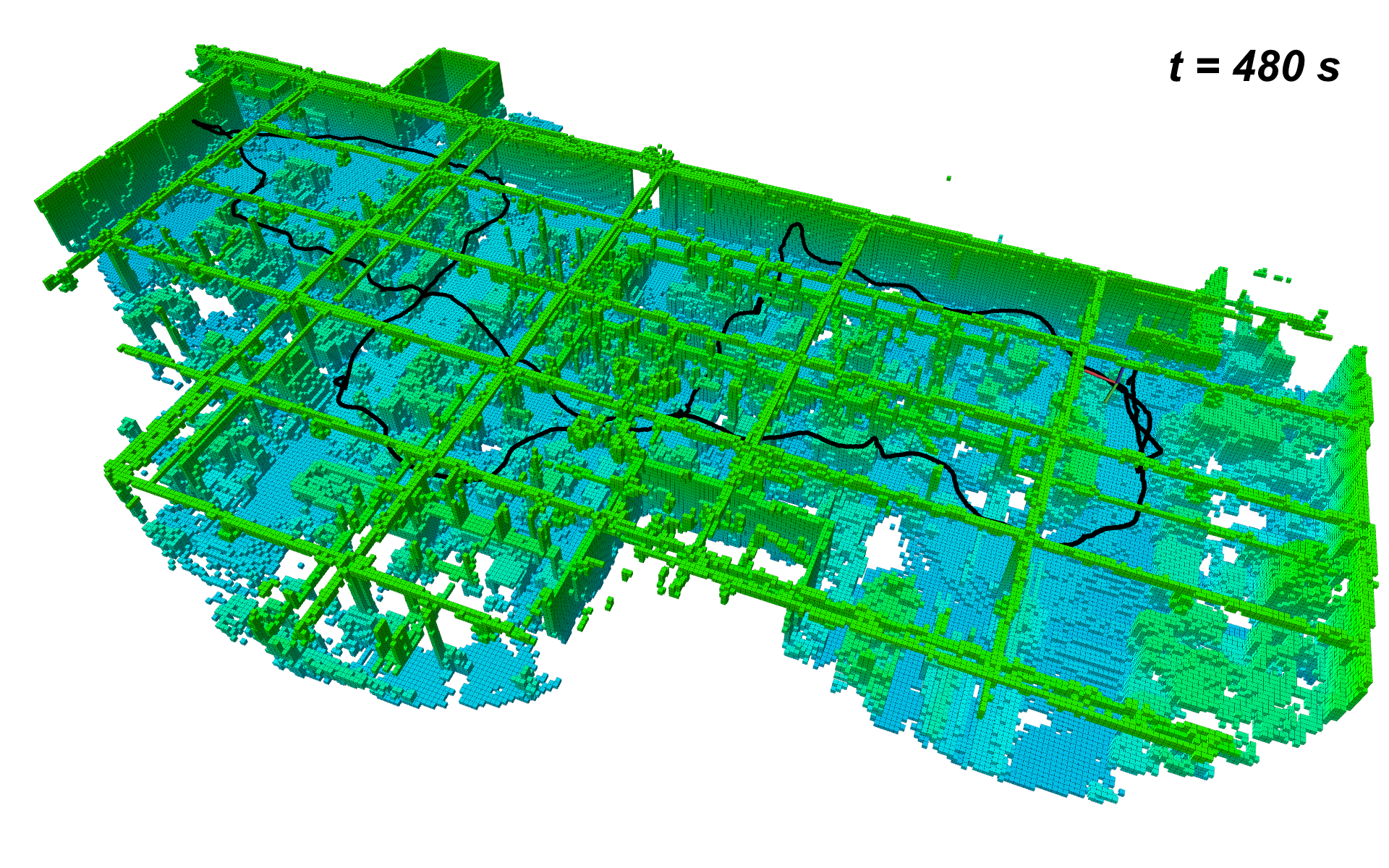}
		\includegraphics[width=.32\textwidth]{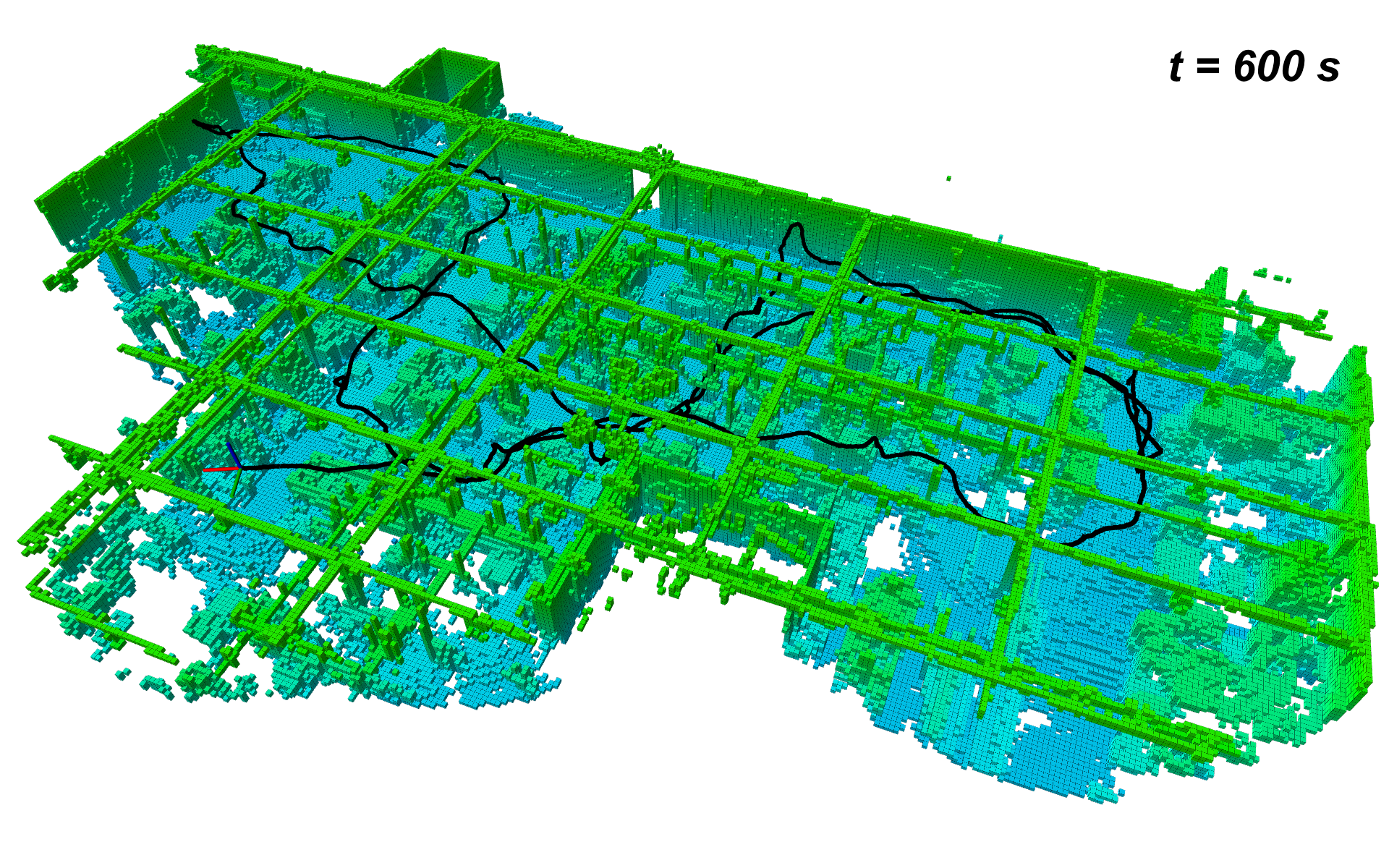}
		\caption{Time progression of an exploration mission in a unstructured warehouse environment. For explicit visualization of the occupancy grid, the OctoMap with 0.2 m leaf size is shown. The ceiling has been removed to provide visibility of the building interior.}
		\label{fig:GeoTech_TEST3_Timelapse}
\end{figure*}

\begin{figure}[h!]
		\centering
		\includegraphics[width=1.0\linewidth]{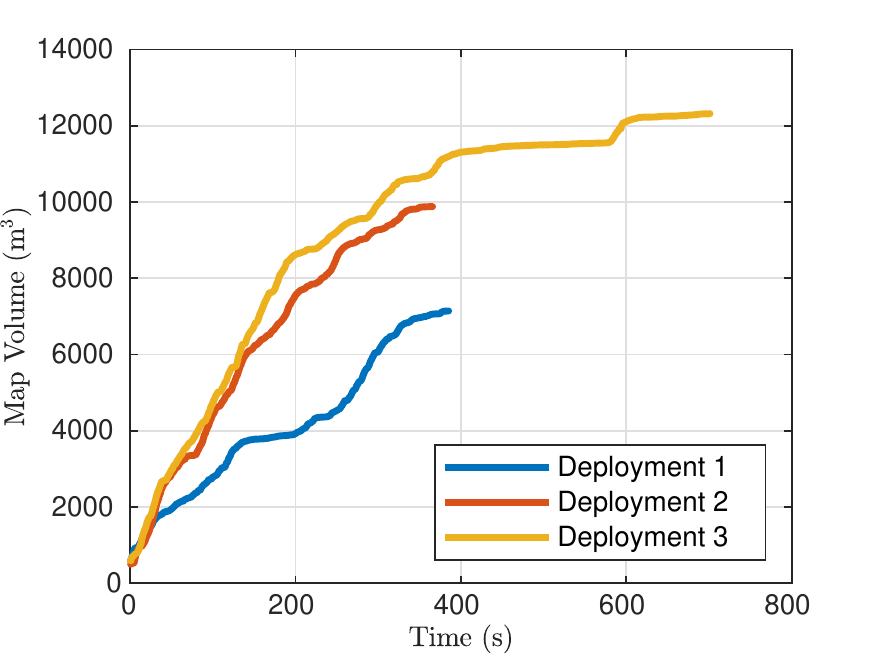}
		\caption{Volume explored by the UAV as a function of time for all three deployments.}
		\label{fig:Time_vs_Volume_Plot}
\end{figure}

This demanding environment has provided an opportunity to evaluate the full proposed solution: frontier-based exploration with potential-field based local control. Given the incredible size of the space, three independent flight tests were conducted in various regions of the warehouse, the results of which are shown in Fig. \ref{fig:GEOTECH_TEST1_TEST2_TEST3_WITH_INSETS}. Clutter in the environment creates occlusions, requiring the agent to precisely modulate position and heading within confined areas to reach its goal pose and continue exploration. A large vertical maneuver is highlighted in Fig. \ref{fig:GEOTECH_TEST1_TEST2_TEST3_WITH_INSETS}e, where the vehicle passes through a narrow slot before flying up towards an unexplored region. With many thin obstacles in the environment, it is likely that some will go undetected during map generation and path planning. Two instances of the local controller recovering the agent from near-collisions are shown in Fig. \ref{fig:GEOTECH_TEST1_TEST2_TEST3_WITH_INSETS}d and Fig. \ref{fig:GEOTECH_TEST1_TEST2_TEST3_WITH_INSETS}f. To provide a sense of the speed of exploration, Fig. \ref{fig:GeoTech_TEST3_Timelapse} captures the map and trajectory at different moments during the third mission. For a more quantitative understanding of the rate of exploration, Fig. \ref{fig:Time_vs_Volume_Plot}, presents the volume of map explored as a function of mission time. \par

These results shared herein have provided validation that the proposed solution is capable of exploring real-world environments that are highly cluttered and unstructured. The interplay between the frontier-based exploration and the potential-based control is demonstrated to be satisfactory in preventing collisions with a variety of obstacles and enabling continued exploration of confined spaces.

\section{Conclusion}
\label{sec:conclusion}
The paper proposed a two-tier planning approach for long-range UAV navigation in highly complex indoor environments. An efficient traversal towards the unexplored map frontiers is achieved using fast marching as a way to perform multi-goal cost-to-go calculations over a Euclidean Signed Distance Field (ESDF) that is generated incrementally on the robot using Voxblox. In many indoor cluttered environments, a single type of perception scheme is not necessarily sufficient to react to the versatile set of obstacles under uncertainties. In order to ensure reliable and safe UAV operations in such scenarios, we extended the Artificial Potential Function (APF) approach to the instantaneous and high resolution depth images to function as the path following controller. A thorough evaluation of the planning architecture is performed through real-world field deployments in a structurally complex warehouse where robustness of the perception and planning solution is of paramount importance along with the exploration efficiency due to extreme levels of clutter. \par
A video of the field deployments is posted at \url{https://youtu.be/UKipgkjVM4k}.

\section*{Acknowledgment}
This work was supported through the DARPA Subterranean Challenge, cooperative agreement number HR0011-18-2-0043. The authors would like to thank Jeff Popiel, the CEO of the Geotech Environmental Equipment, for entertaining our request to conduct autonomous deployments at the warehouse. Also, we would like to thank Zoe Turin from the Bio-Inspired Perception and Robotics Lab at CU Boulder for her help related to the hardware and experiments. 

\bibliographystyle{./bibliography/IEEEtran}
\bibliography{./bibliography/references}

\end{document}